\documentclass{aims} 
\usepackage{amsmath}
\usepackage{paralist}
\usepackage{epsfig} 
\usepackage{graphicx}  
\usepackage{epstopdf} 
\usepackage[colorlinks=true]{hyperref}
\hypersetup{urlcolor=blue, citecolor=red}
\allowdisplaybreaks

\usepackage{amsmath,amssymb,amsthm}
\usepackage{tikz,pgfplots}
\usepackage[toc,page]{appendix}
  \pgfplotsset{compat=newest}
  \usetikzlibrary{shapes,arrows}
  \tikzstyle{neuron} = [circle, draw=white, fill=blue!20!white, minimum height=0.02\textwidth, inner sep=0pt]
  \tikzstyle{line} = [draw, -latex']
\usepackage[margin=2.5cm]{geometry}
\usepackage{hyperref}
\hypersetup{
    colorlinks=false}
\usepackage{verbatim}
\usepackage{calc}  
\usepackage{enumitem} 

\def\currentvolume{X}
 \def\currentissue{X}
  \def\currentyear{200X}
   \def\currentmonth{XX}
    \def\ppages{X--XX}
     \def\DOI{10.3934/xx.xxxxxxx}

\newtheorem{assumption}{Assumption}
\newtheorem{theorem}{Theorem}[section]

\newtheorem{lemma}[theorem]{Lemma}
\newtheorem{proposition}{Proposition}
\newtheorem{conjecture}{Conjecture}

\theoremstyle{definition}
\newtheorem{definition}[theorem]{Definition}
\newtheorem{remark}{Remark}
\newtheorem*{notation}{Notation}

\newcommand{\T}{{\sf T}}

\def\dt#1{\frac{d #1}{d t}}

\DeclareMathOperator{\tr}{tr}

\DeclareMathOperator{\rank}{rank}

\DeclareMathOperator{\Crit}{Crit}

\definecolor{RED}{rgb}{0.7,0,0}
\definecolor{BLUE}{rgb}{0,0,0.69}
\definecolor{GREEN}{rgb}{0,0.6,0}
\definecolor{PURPLE}{rgb}{0.69,0,0.8}

\newcommand{\RED}{\color[rgb]{0.70,0,0}}

\newcommand{\GREEN}{\color[rgb]{0,0.6,0}}

\title[A Geometric Approach of GD Algorithms in Linear NNs]
{A Geometric Approach of Gradient Descent Algorithms in Linear Neural Networks} 

\author[Yacine Chitour and Zhenyu Liao and Romain Couillet]{}


 \email{yacine.chitour@l2s.centralesupelec.fr}
 \email{zhenyu\_liao@hust.edu.cn}
 \email{romain.couillet@univ-grenoble-alpes.fr}

\thanks{$^*$Corresponding author: Yacine Chitour}

\begin{document}


\begin{abstract}
In this paper, we propose a geometric framework to analyze the convergence properties of gradient descent trajectories in the context of linear neural networks. 
We translate a well-known empirical observation of linear neural nets into a conjecture that we call the \emph{overfitting conjecture} which states that, for almost all training data and initial conditions, the trajectory of the corresponding gradient descent system converges to a global minimum. 
This would imply that the solution achieved by vanilla gradient descent algorithms is equivalent to that of the least-squares estimation, for linear neural networks of an arbitrary number of hidden layers. 
Built upon a key invariance property induced by the network structure, we first establish convergence of gradient descent trajectories to critical points of the square loss function in the case of linear networks of arbitrary depth. 
Our second result is the proof of the \emph{overfitting conjecture} in the case of single-hidden-layer linear networks with an argument based on the notion of normal hyperbolicity and under a generic property on the training data (i.e., holding for almost all training data). 
\end{abstract}

\maketitle

\centerline{\scshape Yacine Chitour$^*$}
\medskip
{\footnotesize
 \centerline{Laboratoire des Signaux et Syst{\`e}mes, CentraleSup{\'e}lec, Universit{\'e} Paris-Saclay, France}
} 

\medskip

\centerline{\scshape Zhenyu Liao and Romain Couillet}
\medskip
{\footnotesize
 \centerline{EIC, Huazhong University of Science and Technology, Wuhan, China.}
 \centerline{and}
   \centerline{University Grenoble Alpes, Inria, CNRS, Grenoble INP, LIG, 38000 Grenoble, France.}
}

\bigskip

 \centerline{(Communicated by the associate editor name)}


\section{Introduction}
\label{sec:intro}

Despite the rapid growing list of successful applications of neural networks trained with gradient-based methods in various fields from computer vision \cite{krizhevsky2012imagenet} to speech recognition \cite{mohamed2012acoustic} and natural language processing \cite{collobert2008unified}, our theoretical understanding on these elaborate systems is developing at a more modest pace.

One of the major difficulties in the design of neural networks lie in the fact that, to obtain networks with greater expressive power, one needs to cascade more and more layers to make them ``deeper'' and hope to extract more ``abstract'' features from the (numerous) training data. Nonetheless, from an optimization viewpoint, this ``deeper'' structure often gives rise to non-convex objective functions and makes the optimization dynamics seemingly intractable. In general, finding a global minimum of a generic non-convex function is an NP-complete problem \cite{murty1987some} which is the case for neural networks as shown in \cite{blum1989training} on a very simple network.

Yet, many non-convex problems such as phase retrieval, independent component analysis and orthogonal tensor decomposition obey two important properties \cite{sun2015nonconvex}: i) all local minima are also global; and ii) around any saddle point the objective function has a negative directional curvature (therefore implying the possibility to continue to descend, also referred to as ``strict'' saddles \cite{lee2019first}) and thus allow for the possibility to find global minima with simple gradient descent algorithm. In this regard, the loss geometry of deep neural networks are receiving an unprecedented research interest: in the pioneering work of Baldi and Hornik \cite{baldi1989neural} the landscape of mean square loss was studied in the case of linear single-hidden-layer auto-encoders (i.e., the same dimension for input data and output targets); more recently in the work of Saxe~et~al.\@~\cite{saxe2013exact} the dynamics of the corresponding gradient descent system was first studied, by assuming the input data $X$ empirical correlation matrix $X X^\T $ to be identity, in a linear deep neural network. Then in \cite{kawaguchi2016deep} the author proved that under some appropriate rank condition on the (cascading) weight matrix product, all critical points of a deep linear neural networks are either global minima or saddle points with Hessian admitting eigenvalues with different signs, meaning that linear deep networks are somehow ``close'' to those examples mentioned at the beginning of this paragraph. Nonetheless, the results in \cite{saxe2013exact,kawaguchi2016deep} are incomplete in several ways. First of all, there is no indication that gradient descent trajectories are defined for all time (we did not find an argument of such a fact in the literature) and, even if the aforementioned crucial properties hold true for linear neural networks, they are not sufficient to provide enough (global) information regarding when and how can gradient descent trajectories result in these global minima. Recall that, to escape from saddle points one  generally either chooses the step size depending on \emph{a priori} bounds on the norm of Hessian, thus assuming the gradient descent trajectory converges to a ``well-known'' critical point \cite{lee2019first,panageas2017gradient}, or artificially perturbs the gradient with noise as in \cite{jin2017escape}, or considers networks with some very particular structural properties, e.g., the so-called ``over-parameterization'' which demands in general the widths of the network to be much larger than those of practical interest \cite{allen2018convergence,du2018gradient}. 

\medskip

We attempt to answer the following question: what are the conditions on the initializations and on the  training data for trajectories of simple gradient descent methods to converge with ``high probability'' to global minima in deep linear networks? Let us formulate more precisely the previous question, by proposing a conjecture that we call \emph{overfitting conjecture} (OVF) which says the following: for every positive integer $H$ and almost every choice of data-target pair $(X,Y)$, consider the corresponding continuous time gradient descent algorithm in a $H$-layer linear neural network. 
Then, for almost every initial condition, the gradient descent trajectory defined in Definition~\ref{def:GDD} converges to a global minimum. Here, ``almost every'' refers to the associated Lebesgue measure. The conjecture is referred to as ``overfitting'' since the resulting global minima indeed correspond to networks that are equivalent to the least square solution, explicitly given by 
\[
  W_{LS} = YX^\T (X X^\T)^{-1}
\]
for a given training data-target pair $(X,Y)$ of size $m$, i.e., both $X$ and $Y$ have $m$ columns and $X$ with full row rank. The least square solution is known to suffer from overfitting problem in many cases~\cite{harrell2015regression}; i.e., to provide unsatisfactory performance on unseen data $(x_{new}, y_{new})$, in the sense that $\| y_{new} - W_{LS} x_{new} \|^2$ is much larger than $\frac1m \| Y - W_{LS} X\|_F^2$.





\bigskip

In this paper, we elaborate on the model from \cite{saxe2013exact,kawaguchi2016deep} and evaluate the dynamics of the associated gradient system in a ``continuous'' manner. We first propose a general framework for the geometric understanding of linear neural networks. Based on a cornerstone of ``invariance'' in the parameter space (i.e., the space the network weights) induced by the network cascading structure, we prove the existence for all time of trajectories associated with gradient descent algorithms in linear networks with an arbitrary number of layers and the convergence of these trajectories to critical points of the loss function. The latter result is obtained by first proving that every aforementioned trajectory is bounded and then one gets the convergence with Lojasiewicz's theorem for analytic differential equations, \cite{lojasiewicz1982trajectoires}. We also prove the exponential convergence of trajectories under an extra condition on their initializations. 

By further analyzing the set of critical points, we provide a global picture of the linear gradient descent dynamics. In particular, we characterize, under a generic condition  on the training data, the (finite) set of critical values, i.e., the set of values taken by the loss function on the set of critical points. We also provide a condition on a critical point ensuring that the Hessian there admits at least one negative eigenvalue. This condition is weaker than that proposed in \cite{kawaguchi2016deep}. Moreover, our analysis of second order conditions for the loss function (i.e., linearization of the gradient algorithms) relies on the sole manipulation of the quadratic form associated with the Hessian matrix at a critical point and therefore is more flexible than the similar analysis performed in \cite{kawaguchi2016deep}, where it is based on the use of the Hessian matrix itself. 
 
Finally, we prove the conjecture (OVF) in the case of single-hidden-layer 
linear network by considering the dynamical system defined in Definition~\ref{def:GDD}. 
The argument goes in three steps. In the first one, we carefully analyze the critical set, which is stratified by a finite number of embedded differential manifolds, by showing in particular that, at each point $w$ of such a manifold $M$, the tangent space $T_{w}M$ is exactly equal to the kernel of the Hessian at $w$ of the loss function.
We can then provide a description of the stable manifolds to the dynamics in a compact neighborhood of $w$, thanks to powerful results on normal hyperbolicity, \cite{HPS,Pesin}. The last step of the argument consists in a reasoning by contradiction, which combines the above local description of the dynamics with the behavior of the loss function in that neighborhood.

In particular, our approach improves \cite{du2018algorithmic} in that (i) we  establish the global convergence to critical points of gradient flows for any \emph{deep} linear neural network of depth $H \geq 1$ (while in \cite{du2018algorithmic} only the case of $H =1$ is considered, however in the discrete setting that is of more practical interest) and (ii) while the OVF in the case of $H=1$ has been addressed in \cite{du2018algorithmic}, their approach relies on a specific initialization scheme that is sufficiently small, and in fact within bounded neighborhoods of global minima. On the other hand, our proof does not rely on such initialization and holds, as stated above, for almost every initialization in the state/parameter space. In this sense, our result is \emph{different} and a \emph{good complement} to \cite{du2018algorithmic} in the sense that our analysis is \emph{continuous} but \emph{global}, while the analysis in \cite{du2018algorithmic} is \emph{discrete} but only \emph{local}.

\smallskip
 
As regards linear networks with more than one hidden layer, the above mentioned analysis of critical points is clearly more involved due to the existence of Hessian without negative eigenvalue. As a consequence, a more refined (global) analysis on the basin of attraction of saddle points is required, to attack a proof (or a disproof) of the conjecture (OVF). It would be also interesting to address similar questions (existence, convergence and analysis of the union of basins of attractions) for nonlinear networks. We believe that a starting point would be to come up with enough ``invariants'' along the corresponding trajectories. In \cite{du2018algorithmic}, the authors pointed out that for both the ReLU function $x\mapsto\max(x,0)$ and the Leaky ReLU function $x\mapsto\max(x,\alpha x)$, $0<\alpha<1$, a similar but weaker invariance exists, when one considers the generalized Clarke sub-differential \cite{clarke2008nonsmooth} of the system, signifying a possible loss of control over the non-diagonal entries of the matrices of interest. As such, we are not yet able to conclude for instance boundedness of the trajectories.  

\bigskip

\emph{Structure of the paper}. 
In Section~\ref{sec:system-and-main}, we define the main problem, introduce the notations used throughout the paper as well as a first reduction of the problem. We also provide the invariants along the trajectories and prove convergence results. Section~\ref{sec:characterization-of-critical-points} gathers the analysis of the critical points in a general setting, the main properties on the critical values and the condition for having at least one negative eigenvalue for the Hessian matrix at a critical point. In Section~\ref{sec:proofOVC}, we give the complete argument showing that the conjecture (OVF) holds true for single-hidden-layer linear networks. We finish the paper with conclusion and future perspectives.

\medskip

\emph{Notations}: We denote $\dt{(\cdot)}$ the time derivative, $(\cdot)^\T$ the transpose operator, $\ker(\cdot)$ the kernel of a linear map, i.e., $\ker(M):= \{x, Mx = 0\}$ while $\rank(\cdot)$ and $\tr(\cdot)$ stand the rank and trace operators, respectively. 

\section{System Model and Main Results}
\label{sec:system-and-main}

\subsection{Problem Setup}

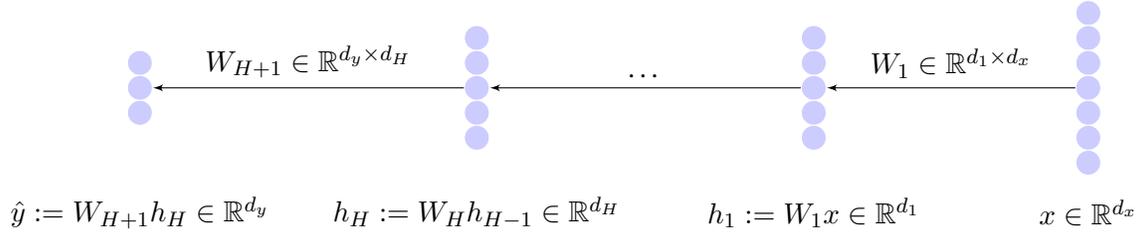
\begin{figure}[htb]
\centering
\begin{tikzpicture}[node distance = 0.02\textwidth, auto]
    \node [neuron] (neuron 11) {};
    \node [neuron, below of=neuron 11] (neuron 12) {};
    \node [neuron, below of=neuron 12] (neuron 13) {};
    \node [neuron, below of=neuron 13] (neuron 14) {};
    \node [neuron, below of=neuron 14] (neuron 15) {};
    \node [neuron, below of=neuron 15] (neuron 16) {};
    \node [neuron, below of=neuron 16] (neuron 17) {};
    \node [below of=neuron 14, yshift=-0.08\textwidth]{$x\in \mathbb{R}^{d_x}$};
    \node [neuron, left of=neuron 11, xshift=-0.2\textwidth, yshift=-0.02\textwidth] (neuron 21) {};
    \node [neuron, below of=neuron 21] (neuron 22) {};
    \node [neuron, below of=neuron 22] (neuron 23) {};
    \node [neuron, below of=neuron 23] (neuron 24) {};
    \node [neuron, below of=neuron 24] (neuron 25) {};
    \node [below of=neuron 23, yshift=-0.08\textwidth]{$h_1 := W_1 x \in \mathbb{R}^{d_1}$};
    \node [neuron, left of=neuron 21, xshift=-0.28\textwidth] (neuron 31) {};
    \node [neuron, below of=neuron 31] (neuron 32) {};
    \node [neuron, below of=neuron 32] (neuron 33) {};
    \node [neuron, below of=neuron 33] (neuron 34) {};
    \node [neuron, below of=neuron 34] (neuron 35) {};
    \node [below of=neuron 33, yshift=-0.08\textwidth]{$h_H:= W_H h_{H-1} \in \mathbb{R}^{d_H}$};
    \node [neuron, left of=neuron 33, xshift=-0.28\textwidth, yshift=0.02\textwidth] (neuron 41) {};
    \node [neuron, below of=neuron 41] (neuron 42) {};
    \node [neuron, below of=neuron 42] (neuron 43) {};
    \node [below of=neuron 42, yshift=-0.08\textwidth]{$\hat y := W_{H+1} h_H \in \mathbb{R}^{d_y}$};
    \path [line] (neuron 14) -- node[above] {$W_1 \in \mathbb{R}^{d_1 \times d_x}$} (neuron 23);
    \path [line] (neuron 23) -- node[above] {\ldots} (neuron 33);
    \path [line] (neuron 33) -- node[above] {$W_{H+1} \in \mathbb{R}^{d_y \times d_H}$} (neuron 42);
\end{tikzpicture}
\caption{Illustration of a $H$-hidden-layer linear neural network}\label{fig:BP-H-layer}
\end{figure}

We start with a linear neural network with $H$ hidden layers as illustrated in Figure~\ref{fig:BP-H-layer}. To begin with, the network structure as well as associated notations are presented as follows.

Let the pair $(X,Y)$ denotes the training data and associated targets, with $X = \begin{bmatrix} x_1, \ldots, x_m \end{bmatrix} \in \mathbb{R}^{d_x \times m}$ and $Y = \begin{bmatrix} y_1, \ldots, y_m \end{bmatrix} \in \mathbb{R}^{d_y \times m}$, where $m$ denotes the number of instances in the training set and $d_x, d_y$ the dimensions of data and targets, respectively. We denote $W_i \in \mathbb{R}^{d_i \times d_{i-1}}$ the weight matrix that connects $h_{i-1}$ to $h_i$ for $i=1,\ldots, H+1$ and set $h_0 = x$, $h_{H+1} = \hat y$ as in Figure~\ref{fig:BP-H-layer}. The network output  for the training set $X$ is therefore given by
\[
  \hat Y = W_{H+1} \ldots W_1 X.
\]
We further denote $W$ the $(H+1)$-tuple of $(W_1,\ldots,W_{H+1})$ for simplicity and work on the mean square error $\mathcal{L}(W)$ given by the following Frobenius norm,
\begin{equation}\label{eq:deep-loss-function-pre}
  \mathcal{L} (W) = \frac12 \| Y - \hat Y \|^2_F = \frac12 \left\| Y - W_{H+1} \ldots W_1 X \right\|_F^2.
\end{equation}
We assume in the sequel that the following assumptions hold true.
\begin{assumption}[Dimension Condition]
\label{ass:deep-dimension-condition}
$m \ge d_x \ge d_y$ and 
$\min(d_1,\ldots,d_H) \ge d_y$.
\end{assumption}

\begin{assumption}[Full Rank Data and Targets]
\label{ass:full-rank-data}
The matrices $X$ and $Y$ are of full (row) rank, i.e., of rank $d_x$ and $d_y$, respectively.
\end{assumption}
\begin{remark}
Assumption~\ref{ass:deep-dimension-condition}~and~\ref{ass:full-rank-data} on the dimension and rank of the training data are realistic and practically easy to satisfy, as discussed in previous works \cite{baldi1989neural,kawaguchi2016deep}. Assumption~\ref{ass:deep-dimension-condition} is demanded here for convenience and our results can be extended to handle more elaborate dimension settings. Indeed, the last part of Assumption~\ref{ass:deep-dimension-condition} is required if one wants to reach the value zero for the effective loss function $L$ to be defined. Similarly, when the training data is rank deficient, the learning problem can be reduced to a lower dimensional one by removing these non-informative (linearly dependent) data in such a way that Assumption~\ref{ass:full-rank-data} holds.
\end{remark}

Under Assumption~\ref{ass:deep-dimension-condition}~and~\ref{ass:full-rank-data}, by performing a singular value decomposition (SVD) on $X$, we obtain
\begin{equation}\label{eq:SVD-X}
    X = U_X \Sigma_X V_X^\T, \quad V_X = \left[\begin{array}{c|c} V_X^1 & V_X^2 \end{array}\right],\qquad \Sigma_X =\left[\begin{array}{c|c} S_X & 0 \end{array}\right], \quad V_X^1 \in \mathbb{R}^{m \times d_x}
\end{equation}
for diagonal and positive $S_X \in \mathbb{R}^{d_x \times d_x}$ so that 
\[
   \mathcal{L}(W) = \frac12 \| Y V_X^1 - W_{H+1} \ldots W_1 U_X S_X \|_F^2 + \frac12 \| Y V_X^2 \|_F^2 := L(W) + \frac12 \| Y V_X^2 \|_F^2
\]
with $W$ the $(H+1)$-tuples of $W_i$ for $i=1,\ldots,H+1$.
By similarly expanding the \emph{effective} loss of the network $L(W)$ according to (the singular value of) the associated reduced target $\bar Y := Y V_X^1$ with $\bar Y = U_Y \Sigma_Y V_Y^\T$, $ \Sigma_Y = \begin{bmatrix} S_Y & 0 \end{bmatrix} \in \mathbb{R}^{d_y \times d_x}$ for diagonal and positive $S_Y \in \mathbb{R}^{d_y \times d_y}$ so that
\begin{equation}\label{eq:deep-loss-function}
  L(W) = \frac12 \|\Sigma_Y - \bar W_{H+1} \bar W_H \ldots \bar W_2 \bar W_1 \|^2_F,
\end{equation}
where we denote $\bar W_1 := W_1 U_X S_X V_Y$, $\bar  W_{H+1} := U_Y^\T W_{H+1}$ and $\bar W_i = W_i$ for $i=2,\ldots,H$. Therefore the state \footnote{The network (weight) parameters $W$ as well as $\bar  W$ evolve through time and are considered to be \emph{state variables} of the dynamical system, while the pair $(X,Y)$ is fixed and thus referred as the ``parameters'' of the given system.} will be $\bar W:=(\bar W_{H+1}, \ldots, \bar W_1)$, with state space equal to
\begin{equation}
    \mathcal{X} = \mathbb{R}^{d_y \times d_H} \times \ldots \times \mathbb{R}^{d_1 \times d_x}
\end{equation} 
and associated metric the Frobenius norm $\tr(\bar W^\T \bar W)=\sum_{i=1}^{H+1}\tr(\bar W_i^\T \bar W_i)$. 

Moreover, for the rest of the paper we will simply use $W$ to denote the state $\bar W$.

With the above notations, we demand in addition the following assumption on the target $\bar Y$.
\begin{assumption}[Distinct Eigenvalues]
The $d_y\times d_y$ matrix $\bar Y\bar Y^\T$ has distinct positive eigenvalues.
\label{ass:distinct-singular-values}
\end{assumption}

Similar to Assumptions~\ref{ass:deep-dimension-condition}~and~\ref{ass:full-rank-data}, Assumption~\ref{ass:distinct-singular-values} is a classical assumption that is demanded in previous works \cite{baldi1989neural,kawaguchi2016deep} and actually holds for an open and dense subset of $\mathbb{R}^{d_y \times m}$.

The objective of this article is to study the gradient descent dynamics (GDD) defined as

\begin{definition}[GDD]\label{def:GDD}
The Gradient Descent Dynamics of $L$ is the dynamical system defined on $\mathcal{X}$ by
\[
 (GDD)\quad \dt{W} = - \nabla_W L,
\]
where $\nabla_w L$ denotes the gradient of the loss function $L$ with respect to $w$, where the gradient is taken according to the choice of the Frobenius norm as metric on the state space. A point $W \in \mathcal{X}$ is a critical point of $L$ if and only if $\nabla_W L = 0$ and we denote $\Crit(L)$ the set of critical points.
\end{definition}

To facilitate further discussion, we drop the bars on $W_i(t)$'s and sometimes the argument $t$ and introduce the following notations.

\begin{notation}\label{notations}
For $1\leq j\leq H+1$, we consider the weight matrix $W_j$ and the corresponding variation $w_j$ of the same size. For simplicity, we denote $W$ and $w$ the $(H+1)$-tuples of $W_j$ and $w_j$, respectively. For two indices $1\leq j,k\leq H+1$, we use $(\Pi W)_j^k$ to denote the product $W_k\ldots W_j$ if $k\geq j$ and identity matrices with appropriate dimension if $k<j$ so that the whole product writes $(\Pi W)_1^{H+1}=W_{H+1}\ldots W_1$ and consequently
\[
  L(W)= \frac12 \|\Sigma_Y -(\Pi W)_1^{H+1} \|^2_F.
\]
For $0\leq r\leq H+1$, and $1\leq j_1<\ldots<j_r$, we use $P^0(W)$ and $P^r_{j_1,\ldots,j_r}(W,w)$ if $r\geq 1$ to denote the following products
\begin{align*}
P^0(W)&=(\Pi W)_1^{H+1},\\
P^r_{j_1,\ldots,j_r}(W,w)&=(\Pi W)_{j_r+1}^{H+1}w_{j_r}(\Pi W)_{j_{r-1}+1}^{j_r-1}w_{j_{r-1}}\ldots (\Pi W)_{j_1+1}^{j_2-1}w_{j_1}(\Pi W)_1^{j_1-1}.
\end{align*}
For instance, we have, for $1\leq j<k\leq H+1$,
\begin{align*}
P^1_j(W,w)&=(\Pi W)_{j+1}^{H+1}w_{j}(\Pi W)_{1}^{j-1},\\
P^2_{j,k}(W,w)&=(\Pi W)_{k+1}^{H+1}w_{k}(\Pi W)_{j+1}^{k-1}w_{j}(\Pi W)_{1}^{j-1}.
\end{align*}
\end{notation}

We can use the above notations to derive the first-order variation of the loss function $L$ and hence the GDD equations. To this end, set
\begin{equation}\label{eq:MML}
  M=\Sigma_Y -(\Pi W)_1^{H+1},
\end{equation}
so that
\begin{align*}
  L(W+w)&=L(W)-\sum_{j=1}^{H+1}\tr(P^1_j(W,w)M^\T)+O(\|w\|^2),\\
  &=L(W)-\sum_{j=1}^{H+1}\tr(w_{j}(\Pi W)_{1}^{j-1}M^\T(\Pi W)_{j+1}^{H+1})
  +O(\|w\|^2),
\end{align*}
where $O(\|w\|^2)$ stands for polynomial terms of order equal or larger than two in the $w_j$'s. We thus obtain, for $1\leq j\leq H+1$,
\begin{equation}\label{eq:deep-linear-gradient}
\dt{ W_j} = \left[(\Pi W)_{j+1}^{H+1}\right]^\T M\left[(\Pi W)_{1}^{j-1}\right]^\T.
\end{equation}

\subsection{Convergence Analysis}
We start with the existence for all $t \ge 0$ of all gradient descent trajectories, based on which we then establish their global convergence to critical points. While one expects the gradient descent algorithm to converge to critical points, this may not always be the case. Two possible (undesirable) situations are 1) a trajectory is unbounded or 2) it oscillates ``around'' several critical points without convergence, i.e., along an \( \omega\)-limit set made of a continuum of critical points (see \cite{teschl2012ordinary} for notions on \( \omega\)-limit sets). The property of an iterative algorithm (like gradient descent) to converge to a critical point for any initialization is referred to as ``global convergence'' \cite{zangwill1969convergence}.  However, it is very important to stress the fact that it does not imply (contrary to what the name might suggest) convergence to a global minimum for all initializations.

To answer the convergence question, we resort to Lojasiewicz's theorem for the convergence of a gradient descent flow of the type of \eqref{eq:deep-linear-gradient} with real analytic right-hand side, \cite{lojasiewicz1982trajectoires}, as formally recalled below.
\begin{theorem}[Lojasiewicz's theorem, \cite{lojasiewicz1982trajectoires}]
Let \( L \) be a real analytic function and let \(W(\cdot) \) be a solution trajectory of the gradient system given by Definition~\ref{def:GDD} such that \( \sup_{t \ge 0} \| W(t) \| < \infty \), i.e. \(W(\cdot) \)  is bounded. Then \(W(\cdot)\) converges to a critical point of \(L\), as \( t \to \infty \). The rate of convergence is determined by the associated Lojasiewicz exponent \cite{d2005explicit}.
\label{theo:lojasiewicz}
\end{theorem}

\begin{remark}
\label{rem:extension-lojasiewicz}
Since the fundamental (strict) gradient descent direction (as in Definition~\ref{def:GDD}) in Lojasiewicz's theorem can in fact be relaxed to a more general angle condition (see for example Theorem~2.2 in~\cite{absil2005convergence}), the line of argument developed in the core of this paper may be similarly followed to prove the global convergence of more advanced optimizers (e.g., SGD, SGD-Momentum~\cite{qian1999momentum}, ADAM~\cite{kingma2014adam}, etc.), for which the direction of descent is not strictly the opposite of the gradient direction. This constitutes an important direction of future exploration.
\end{remark}

Since the loss function \(L\) is a polynomial of degree \( (H+1)^2 \) in the components of \(W\), Lojasiewicz's theorem ensures that if a given trajectory of the gradient descent flow is bounded (i.e., it remains in a compact set for every \(t \ge 0\)) it must converge to a critical point with a guaranteed rate of convergence. In particular, the aforementioned phenomenon of ``oscillation'' cannot occur and we are left to ensure the absence of unbounded trajectories. The following lemma characterizes the ``invariants'' along trajectories of GDD, inspired by \cite{saxe2013exact} which essentially considered the case where all dimensions are equal to one. These invariants will be used at several stages of the paper.

\begin{lemma}[Invariant in GDD]\label{lem:invariant-in-deep-GDD}
Consider any trajectory of the gradient system given by \eqref{eq:deep-linear-gradient}. Then, for $1\leq j\leq H$, the value of $W_{j+1}^\T W_{j+1} - W_j W_j^\T$ remains constant on its interval of definition, i.e.,  
\begin{equation}\label{eq:inv0}
  W_{j+1}^\T W_{j+1} - W_j W_j^\T = \left. (W_{j+1}^\T W_{j+1} - W_j W_j^\T) \right|_{t=0}:=C_j,
\end{equation}
for $1\leq j\leq H$. As a consequence, there exist constant real numbers $c_j$, $1\leq j \leq H$, such that, along a trajectory of the gradient system given by \eqref{eq:deep-linear-gradient}, one has on the interval of definition of the trajectory,
\begin{equation}\label{eq:inv-W}
\Vert W_{j+1}\Vert_F^2=\Vert W_1\Vert_F^2+c_{j+1}.
\end{equation}
\end{lemma}

\begin{proof}
With the above notations together with \eqref{eq:deep-linear-gradient}, one gets
\begin{align*}
  \dt{(W_j W_j^\T)} &= \left[(\Pi W)_{j+1}^{H+1}\right]^\T M\left[(\Pi W)_{1}^{j-1}\right]^\T W_j^\T+
  W_j(\Pi W)_{1}^{j-1}M^\T(\Pi W)_{j+1}^{H+1}\\
  &=\left[(\Pi W)_{j+1}^{H+1}\right]^\T M\left[(\Pi W)_{1}^{j}\right]^\T + (\Pi W)_{1}^{j}M^\T(\Pi W)_{j+1}^{H+1}\\
  &=W_{j+1}^\T\left[(\Pi W)_{j+2}^{H+1}\right]^\T M\left[(\Pi W)_{1}^{j}\right]^\T+ (\Pi W)_{1}^{j}M^\T(\Pi W)_{j+2}^{H+1}W_{j+1}\\
  &= \dt{(W_{j+1}^\T W_{j+1})},
\end{align*}
hence the conclusion of \eqref{eq:inv0}. To deduce \eqref{eq:inv-W}, it remains to add the above equations, up to transposition, from the indices $j$ to $H+1$, and then take the trace. 
\end{proof}

\begin{remark}\label{rem:nonL}
Lemma~\ref{lem:invariant-in-deep-GDD} provides a key structural property of the GDD in linear networks, which is instrumental to ensure the boundedness of the gradient descent trajectories and thus in turn to prove the convergence to critical points. Moreover, similar property holds in more elaborate neural networks, for example Lemma~\ref{lem:invariant-in-deep-GDD} holds for the popular softmax-cross-entropy loss with one-hot vector targets, with and without $\ell_2$ regularization \cite{arora2018optimization}; also the conservation of norms in \eqref{eq:inv-W} holds true in nonlinear neural networks with ReLU and Leaky ReLU nonlinearities \cite{du2018algorithmic}. 
\end{remark}

Based on Lemma~\ref{lem:invariant-in-deep-GDD}, we introduce the following lemma which is the core argument to show that all trajectories of the GDD are indeed bounded.

\begin{lemma}\label{lem:est-conv}
There exists a positive constant $\gamma_0<1$ only depending on $H$ and on the dimensions involved in the problem such that, for every $W\in\mathcal{X}$, there exist two polynomials $P$ and $Q$ of degree at most $H$ with nonnegative coefficients only depending on $H$ and on the matrices $C_j, 1 \leq j \leq H$, defined in \eqref{eq:inv0} so that one has
\begin{align}
\gamma_0\Vert W_{H+1}\Vert_F^{2(H+1)}-P(\Vert W_{H+1}\Vert_F^2) &\leq \tr\big(\left[(\Pi W)_{1}^{H+1}\right]^\T(\Pi W)_{1}^{H+1}\big) \nonumber \\
&\leq \Vert W_{H+1}\Vert_F^{2(H+1)}+Q(\Vert W_{H+1}\Vert_F^2).\label{eq:main-est}
\end{align}
\end{lemma}

\begin{proof}
Lemma~\ref{lem:est-conv} is established by induction on $H$ and hence on the number of factors in the state space $\mathcal X$. In the sequel, the various constants (generically denoted by $K$) are positive and only dependent on the $C_j$'s and the $c_j$'s defined in Lemma~\ref{lem:invariant-in-deep-GDD}, thus independent of $t\geq 0$ in the interval of definition of the trajectory. The case $H=0$ is immediate by taking $P=Q=0$ and any $\gamma_0\in (0,1)$. We assume that it holds for $H$ and treat the case $H+1$. One has
\begin{align*}
\tr\big(\left[(\Pi W)_{1}^{H+1}\right]^\T(\Pi W)_{1}^{H+1}\big)&=\tr\big(W_{1}^\T W_2^\T\ldots W_{H+1}^\T W_{H+1}\ldots W_2W_1\big)\\
&=\tr\big(W_{H+1}W_H\ldots W_2W_1W_{1}^\T W_2^\T\ldots W_{H}^\T W_{H+1}^\T\big).
\end{align*}
Using \eqref{eq:inv0}, we replace the product $W_1W_{1}^\T$ by $W_2^\T W_{2}-C_1$ in the above expression to obtain
\begin{align*}
\tr\big(\left[(\Pi W)_{1}^{H+1}\right]^\T(\Pi W)_{1}^{H+1}\big)&=
\tr\big(W_{H+1}W_H\ldots (W_2W_2^\T)^2\ldots W_{H}^\T W_{H+1}^\T \big)\\
&-\tr\big(W_{H+1}W_H\ldots W_2C_1W_2^\T\ldots W_{H}^\T W_{H+1}^\T \big).
\end{align*}
First note that 
$$
\tr\big(W_{H+1}\ldots W_2C_1W_2^\T\ldots W_{H}^\T W_{H+1}^\T \big)=\tr(AC_1),
$$ 
where $A:=\left[(\Pi W)_{2}^{H+1}\right]^\T(\Pi W)_{2}^{H+1}$ is symmetric and nonnegative definite. By using the fact that 
$$
\vert \tr(AC_1) \vert \leq \Vert C_1\Vert_F \tr(A),
$$
one deduces that 
\begin{align*}
%
%
%
&-\Vert C_1\Vert_F\tr\big(W_{H+1} \ldots W_2W_2^\T\ldots W_{H+1}^\T \big)+
\tr\big(W_{H+1} \ldots (W_2W_2^\T)^2\ldots W_{H+1}^\T \big) \\
&\leq \tr\big(\left[(\Pi W)_{1}^{H+1}\right]^\T(\Pi W)_{1}^{H+1}\big) \\ 
&\leq \Vert C_1\Vert_F\tr\big(W_{H+1} \ldots W_2W_2^\T\ldots W_{H+1}^\T \big) + \tr\big(W_{H+1} \ldots (W_2W_2^\T)^2\ldots W_{H+1}^\T\big).
\end{align*}
We now apply the induction hypothesis for elements $\tilde{W}=(W_{H+1},W_H,\ldots,W_2)$ in the state space $\tilde{{\mathcal X}}=\mathbb{R}^{d_y \times d_H} \times \ldots \times \mathbb{R}^{d_2 \times d_1}$ and, using the estimate \eqref{eq:main-est} corresponding to $A$, one deduces that
\begin{align*}
&-P(\Vert W_{H+1}\Vert_F^2)+\tr\big(W_{H+1}W_H\ldots (W_2W_2^\T)^2\ldots W_{H}^\T W_{H+1}^\T \big) \\
&\leq \tr\big(\left[(\Pi W)_{1}^{H+1}\right]^\T(\Pi W)_{1}^{H+1}\big) \\ 
&\leq\tr\big(W_{H+1}W_H\ldots (W_2W_2^\T)^2\ldots W_{H}^\T W_{H+1}^\T \big)
+Q(\Vert W_{H+1}\Vert_F^2).
\end{align*}
where $P,Q$ are polynomials of degree $H$. Again with \eqref{eq:inv0}, we replace the term $(W_2W_{2}^\T)^2$ by $(W_3^\T W_{3}-C_2)^2$. By developing the square inside the larger product, we obtain as principal term 
\begin{align*}
\tr\big(W_{H+1}\ldots W_3(W_3^\T W_3)^2W_3^\T\ldots W_{H+1}^\T \big)=\tr\big(W_{H+1} \ldots (W_3W_3^\T)^3\ldots W_{H+1}^\T \big),
\end{align*}
with lower order terms upper and lower bounded, thanks to the induction hypothesis, by $Q(\Vert W_{H+1}\Vert_F^2)$ and $-P(\Vert W_{H+1}\Vert_F^2)$, respectively, for some polynomials $P,Q$ of degree $H$ with nonnegative coefficients. We then similarly proceed by replacing the term $(W_3W_{3}^\T)^3$ by $(W_4^\T W_{4}-C_3)^3$ and so on, so as to end up with the following estimate
\begin{align*}
&\tr\big((W_{H+1}W_{H+1}^\T)^{H+1}\big)-P(\Vert W_{H+1}\Vert_F^2) \\
&\leq \tr\big(W_{H+1}W_H\ldots W_3(W_3^\T W_3)^2W_3^\T\ldots W_{H}^\T W_{H+1}^\T \big)\\
&\leq\tr\big((W_{H+1}W_{H+1}^\T)^{H+1}\big)+Q(\Vert W_{H+1}\Vert_F^2),
\end{align*}
for some polynomials $P,Q$ of degree $H$ with nonnegative coefficients. Recall that, for $k,l$ positive integers, there exists a positive constant $\gamma_0<1$ only depending on $k,l$ such that for every $k\times k$ nonnegative symmetric matrix $S$, one has
\[
  \gamma_0\big(\tr(S)\big)^l\leq \tr(S^l)\leq \big(\tr(S)\big)^l.
\]
(Indeed, it is enough to see that for diagonal matrices with non negative coefficients and apply H\"older's inequality.) This concludes the proof of the lemma.
\end{proof}

With Lemma~\ref{lem:est-conv}, we are in position to introduce the main result of this section on the global convergence of every gradient descent trajectory to a critical point.
\begin{proposition}[Global Convergence of GDD to Critical Points]
\label{prop:deep-convergence-to-critical-points}
Let $(X,Y)$ be a data-target pair.
Then, every trajectory of the corresponding gradient flow described by Definition~\ref{def:GDD} converges to a critical point.
\end{proposition}

\begin{proof}
With Lojasiewicz’s theorem, we are left to prove that each trajectory of \eqref{eq:deep-linear-gradient} remains in a compact set. Taking into account  \eqref{eq:inv-W}, it is enough to prove that $\Vert W_{H+1}\Vert_F$ is bounded along each trajectory. To this end, denoting $g:=\Vert W_{H+1}\Vert_F^2 = \tr(W_{H+1}W_{H+1}^\T)$ and considering its time derivative, one gets
\begin{align}
  \dt{g} &= 2\tr\left(M\left[(\Pi W)_{1}^{H+1}\right]^\T\right) \nonumber \\
  &=2\tr\left(\Sigma_Y\left[(\Pi W)_{1}^{H+1}\right]^\T\right)
  -2\tr\big(\left[(\Pi W)_{1}^{H+1}\right]^\T(\Pi W)_{1}^{H+1}\big). \label{eq:g1}
 \end{align}
 For the first term on the right-hand side of \eqref{eq:g1}, we use \eqref{eq:inv-W} to upper bound it by a polynomial in $\Vert W_{H+1}\Vert_F$ of degree $H+1$ and we use \eqref{eq:main-est} to get that
 \[
  \dt{g}\leq -2\gamma_0 g^{H+1}+ K_1(1+g^H),
\]
for some positive constant $K_1$ only depending, as $\gamma_0$
on the initial condition of the trajectory. Clearly, there exists a positive constant $K_2$ depending on the trajectory such that 
the right-hand side of the above trajectory is negative for $g>K_2$. This implies at once that $\limsup g$, as $t$ tends to infinity, is less than or equal to $K_2$, and thus, the trajectory remains in a compact set. This 
concludes the proof of Proposition~\ref{prop:deep-convergence-to-critical-points}. The guaranteed rate of convergence can be obtained from estimates associated with polynomial gradient systems \cite{d2005explicit}.
\end{proof}
In full generality, Lojasiewicz's theorem guarantees that the rate of convergence of trajectories of an analytic gradient system as $t \to \infty$ is only polynomial, i.e., of the type $t^{-\alpha}$, for some $\alpha > 0$. For the GDD in Definition~\ref{def:GDD} and under an additional dimension assumption, we can achieve exponential decay rate for large sets of initial conditions, as given in the next proposition.

\begin{proposition}[Exponential Convergence of GDD]\label{prop:exponential-convergence}
Let Assumption~\ref{ass:deep-dimension-condition} holds and assume in addition that  $d_1 \ge d_2 \ge \ldots \ge d_H\ge d_y$ and the 
initialization $C_j:= \left[W_{j+1}^\T W_{j+1} - W_j W_j^\T\right]_{t=0} \in 
\mathbb{R}^{d_j \times d_j}$ has at least $d_{j+1}$ positive eigenvalues for 
$j=1,\ldots,H$. Then, every trajectory of (GDD) converges to a global 
minimum with an exponential rate equal to the product of all the 
$d_{j+1}$-smallest eigenvalue of $C_j$, $1\leq j\leq H$.
\end{proposition}
\begin{proof}
Under Notations~\ref{notations} and by using \eqref{eq:deep-linear-gradient}, we have a result of independent interest  namely the differential equation satisfied by $M$
\begin{align*}
    \dt{M} &= - \sum_{j=1}^{H+1} (\Pi W)_{j+1}^{H+1} \dt{W_j} (\Pi W)_1^{j-1} \\ 
    &= - \sum_{j=1}^{H+1} (\Pi W)_{j+1}^{H+1} \left[ (\Pi W)_{j+1}^{H+1} \right]^\T M \left[ (\Pi W)_1^{j-1} \right]^\T (\Pi W)_1^{j-1}.
\end{align*}
We deduce the dynamics of the square-norm of $M$ (which is nothing but expressing infinitesimally the decrease of the loss function $L$ along trajectories of (GDD))
\begin{align*}
 \dt{\Vert M\Vert^2_F} &= - 2 \sum_{j=1}^{H+1} \tr \left( M^\T (\Pi W)_{j+1}^{H+1} \left[ (\Pi W)_{j+1}^{H+1} \right]^\T M \left[ (\Pi W)_1^{j-1} \right]^\T (\Pi W)_1^{j-1} \right) \\ 
 &=- 2 \sum_{j=1}^{H+1}\textrm{RHS}_j, 
\end{align*}
with $\textrm{RHS}_j= \tr\Big(\left(\left[(\Pi W)_{j+1}^{H+1} \right]^\T M \left[ (\Pi W)_1^{j-1} \right]^\T\right)^\T \left[(\Pi W)_{j+1}^{H+1} \right]^\T M \left[ (\Pi W)_1^{j-1} \right]^\T\Big)$.
It is immediate to see that $\textrm{RHS}_j$, $1\leq j\leq H$ is indeed non 
negative as the trace of a non negative real symmetric matrix. We next 
rewrite each of them as follows,
$$
\textrm{RHS}_j=\tr\big(W_{j+1}W_{j+1}^\T K_j K_j^\T\big), \hbox{ with }
K_j=\left[(\Pi W)_{j+2}^{H+1} \right]^\T M \left[ (\Pi W)_1^{j-1} \right]^\T.
$$
Recall now that, for any real matrices $A,B$ with the same number of rows, it holds 
$$
\tr( AA^\T BB^\T)\geq \lambda_{\min} (AA^\T) \tr(BB^\T),
$$
where $\lambda_{\min} (\cdot)$ denotes  the minimum eigenvalue of a real 
symmetric non negative matrix. We apply the previous fact to deduce that
$\textrm{RHS}_j\geq \lambda_{\min} (W_{j+1}W_{j+1}^\T)\tr(K_j K_j^\T)$. Handling similarly $\tr(K_j K_j^\T)$ and proceeding recursively, we get
$$
 \dt{\Vert M\Vert^2_F}\leq -2\sum_{j=1}^{H+1} \prod_{k=1}^{j-1} \lambda_{\min} (W_k^\T W_k) \tr \left( M^\T (\Pi W)_{j+1}^{H+1} \left[ (\Pi W)_{j+1}^{H+1} \right]^\T M \right).
 $$
 We further handle the traces in the above equation similarly to the $\textrm{RHS}_j$'s and finally obtain 
\begin{equation}\label{eq:M-dyn1}
 \dt{\Vert M\Vert^2_F}\leq
 - 2 \sum_{j=1}^{H+1} \prod_{k=1}^{j-1} \lambda_{\min} (W_k^\T W_k) \prod_{l=j+1}^{H+1} \lambda_{\min} (W_l W_l^\T)\Vert M\Vert^2_F.
\end{equation}
Since $d_1 \ge \ldots \ge d_j \ge d_{j+1} \ge \ldots \ge d_H$, the rank of the $d_l\times d_l$ matrix $W_l W_l^\T$ is at 
most $d_{l+1}\leq d_l$ for any $1\leq l\leq H+1$. Hence  
$\lambda_{\min} (W_l W_l^\T)=0$ as soon as $d_{l+1}< d_l$ for some index $l$. 
This is why we only focus the term $j=1$ in \eqref{eq:M-dyn1} and get, for every $t\geq 0$, 
\begin{equation}
 \dt{\Vert M\Vert^2_F}\leq -2\alpha(t)\Vert M\Vert^2_F,\hbox{ with }
 \alpha(t)=\prod_{l=2}^{H+1} \lambda_{\min} (W_l W_l^\T)\geq 0,
 \end{equation}
 where we have emphasized the dependence of the exponential rate with respect
 to the time.
   
 Let us now bound from below $\alpha(\cdot)$ by a non negative constant. For that 
 purpose, let use the notation $\lambda_i(A)$ to denote the $i$-th eigenvalue of $A$ arranged in algebraically non-decreasing order so that $\lambda_1 (A) = \lambda_{\min} (A)$, for any real symmetric square matrix $A$.
  
  From Lemma~\ref{lem:invariant-in-deep-GDD} and Weyl's inequality (e.g., \cite[Corollary~4.3.12]{horn1990matrix}), we have, for $l= 1,\ldots,H$ that
\[
  \lambda_{i}(W_{l+1}^\T W_{l+1}) \ge \lambda_{i}(C_l) + \lambda_{\min}(W_l W_l^\T) \ge \lambda_{i}(C_l)
\]
 We next prove that under the assumptions of the proposition that 
 $\alpha$ is bounded below by a non negative constant.
Note that $W_{l+1}^\T W_{l+1}$ is of maximum rank $d_{l+1}$ and thus admits at least $d_l - d_{l+1}$ zero eigenvalues so that
\[
  \lambda_i (W_{l+1}^\T W_{l+1}) = 0, \quad \lambda_i (C_l) \le 0,
\]
for $i = 1, \ldots, d_l - d_{l+1}$. Moreover, since for $i = 1, \ldots, d_{j+1}$ we also have,
\[
  \lambda_{i+ d_l - d_{l+1}} (W_{l+1}^\T W_{l+1}) = \lambda_i (W_{l+1} W_{l+1}^\T)
\]
we obtain at once that for $i = 1, \ldots, d_{l+1}$,
\[
  \lambda_i (W_{l+1} W_{l+1}^\T) = \lambda_{i+ d_l- d_{l+1}} (W_{l+1}^\T W_{l+1}) \ge \lambda_{i+ d_l- d_{l+1}} (C_l).
\]
By taking $j=1$, one deduces from \eqref{eq:M-dyn1} that, for every $t\geq 0$,
\[
  \alpha(t)\ge \alpha:=\prod_{l=1}^H \lambda_{d_l - d_{l+1} + 1} (C_l).
\]
Recalling that $\lambda_{d_l - d_{l+1} + 1} (C_l)$ is the $d_{l+1}$-smallest eigenvalue of $C_l$, one deduces from the assumptions of the proposition that $\alpha>0$, yielding an exponential decay of $M$ to zero.
That result, combined with Proposition~\ref{prop:deep-convergence-to-critical-points},  concludes the proof.
\end{proof}

\begin{remark} 
Assumptions of Proposition~\ref{prop:exponential-convergence} are verified by large sets of initial conditions, once it is noticed that, for $1\leq l\leq H$,
$$
\lambda_{d_l - d_{l+1} + 1} (C_l)\geq \lambda_1(W_{l+1}W_{l+1}^\T)-\lambda_1(W_{l}W_{l}^\T).
$$
Indeed, it is always possible to choose recursively $W_{l+1}$ for a given $W_l$ so that the right-hand side of above inequality is positive.
\end{remark}

\subsection{Conjecture (OVF)}
While in the very specific case of Proposition~\ref{prop:exponential-convergence} where the network is restricted to have a pyramidal structure and satisfy some particular initialization conditions, every trajectory of the GDD is known to converge to a global minimum with $M = 0$ and $L=0$, in more general settings of initializations we have no idea whether the gradient descent will be ``trapped'' in critical points that are not global minima. In this section, we propose a stronger possible behavior on the convergence of the GDD trajectories: we make the conjecture that for almost every initial condition, the corresponding GDD trajectory converges to a global minimum.

\medskip

Recall that in linear networks that every local minimum is global and there is no local maximum, see  \cite{kawaguchi2016deep}) or the next section. Moreover, the basin of attraction of a critical point is the set of initializations for which the GDD trajectories converge to that given critical point.

We also refer here as ''saddle point'' a critical point which is not a local extremum. Hence, concretely in our proposed framework, we focus on the state space $\mathcal{X}$ and first evaluate ``how much'' is occupied by the saddle points: we stratify the set of critical points \(\Crit(L)\) in \(d_y + 1\) subsets, one of them (\(\Crit_{d_y}(L)\)) corresponding to the set of global minima and the \(d_y\) others, \(\Crit_r(L)\) with \(r=0,\ldots,d_y-1\), corresponding to the set of saddle points. 

Moreover, since every trajectory of (GDD) converges to a critical point as a results of Proposition~\ref{prop:deep-convergence-to-critical-points}, one deduces that the union over all the critical points of the basins of attraction associated with each critical points is equal to the state space $\mathcal{X}$. One can therefore formulate the overfitting conjecture (OVF) as follows.

\begin{conjecture}[Conjecture (OVF)]\label{con:OVF}
Let $(X,Y)$ be a data-target pair satisfying Assumptions~\ref{ass:deep-dimension-condition} to \ref{ass:distinct-singular-values}. Then, for almost every initial condition $W_0\in \mathcal{X}$, the trajectory of (GDD) starting at $W_0$ converges to a global minimum.
In other words, the union of the basins of attraction associated with the saddle points of 
$L$ is a set of zero (Lebesgue) measure. 
\end{conjecture}

\begin{remark}[Least square solution]\label{rem:linear-regression}
If we write the objective function in \eqref{eq:deep-loss-function-pre} as $\mathcal{L}(W) = \frac12 \| Y - W X \|_F^2$ by considering the product $W_{H+1} \ldots W_1$ as a single matrix $W$. This optimization problem is then convex and the only optimal $W$ that minimizes $\mathcal{L}$ is the least square solution $W_{LS}$, given explicitly as
\[
  W_{LS} = YX^\T(XX^\T)^{-1}
\]
for invertible $XX^\T$. Despite its simple form, the above least square solution is known to easily over-fit and yields unsatisfactory performance in many cases, see \cite{harrell2015regression}.
\end{remark}

A natural way to address the following conjecture consists in performing 
a study on the local behavior of gradient descent trajectories ``around'' each saddle point, so as to measure its basin of attraction. In the following section we provide a precise characterization of critical points, which, serves as a significant step to prove the conjecture (OVF) in the case $H=1$ under the additional Assumption~\ref{ass:distinct-critical-values} in Section~\ref{sec:proofOVC}.

\section{Study of Critical Points}
\label{sec:characterization-of-critical-points}
In this section, we assume that Assumptions~\ref{ass:deep-dimension-condition}~and
  ~\ref{ass:full-rank-data} hold true.
\subsection{Critical Points Condition}
Decomposing $W_1 = \left[\begin{array}{c|c} W_{1,1} & W_{1,2} \end{array}\right]$ with $W_{1,1} \in \mathbb{R}^{d_1\times d_y}, W_{1,2} \in \mathbb{R}^{d_1 \times (d_x - d_y)}$, the effective loss $L$ writes

\begin{equation}\label{eq:deep-loss-function-simple}
  L(W) = \frac12 \| S_Y - (\Pi W)_2^{H+1} W_{1,1} \|^2_F + \frac12 \| (\Pi W)_2^{H+1} W_{1,2} \|^2_F
\end{equation}
where we recall $(\Pi W)_2^H \in \mathbb{R}^{d_H \times d_1}$ and $W_{H+1} \in \mathbb{R}^{d_y \times d_H}$ so that the product $(\Pi W)_2^{H+1} \in \mathbb{R}^{d_y \times d_1}$. To retrieve information on the  critical points of the loss function $L$ in \eqref{eq:deep-loss-function-simple} as well as their basins of attraction, we shall expand the first two order variations of $L(W + w)$ as described in the following proposition.

\begin{proposition}[Variation of $L$ in Deep Nets]\label{prop:variation-deep}
For $W=(W_{1,1},W_{1,2},W_2,\ldots,W_{H+1})$ and $w=(w_{1,1},w_{1,2},w_2,\ldots,w_{H+1})$ in $\mathcal{X}$,
set $M:=M(W) = S_Y - (\Pi W)_2^{H+1} W_{1,1} \in \mathbb{R}^{d_y \times d_y}$. We then have the following expansion for $L(W + w)$,

\begin{equation}
L(W + w)=L(W) + \Delta_W(w) + H_W(w)  + O \left( \|w\|^3 \right),
\end{equation}
with $L(W) = \frac12 \|M\|_F^2 + \frac12 \| (\Pi W)_2^{H+1} W_{1,2}\|_F^2$ and
\begin{align*}
  \Delta_W(w) &= - \sum_{j=2}^H \tr\left( W_{H+1} Q_j^1 (W,w) W_{1,1} M^\T \right) - \tr \left( w_{H+1} (\Pi W)_2^H W_{1,1} M^\T\right)\\
  & + \sum_{j=2}^H \tr \left( W_{H+1} Q_j^1(W,w) W_{1,2} W_{1,2}^\T [(\Pi W)_2^{H+1} ]^\T \right) \\ 
  &+ \tr \left( w_{H+1} (\Pi W)_2^H W_{1,2} W_{1,2}^\T [(\Pi W)_2^{H+1} ]^\T \right)\\
  & - \tr \left( (\Pi W)_2^{H+1} w_{1,1} M^\T \right) + \tr\left( (\Pi W)_2^{H+1} w_{1,2} W_{1,2}^\T [(\Pi W)_2^{H+1}]^\T \right),
\end{align*}
where we denote (similarly to Notations~\ref{notations}) the following products
\begin{align*}
  Q^0(W) &= (\Pi W)_2^H, \\
  Q_j^1(W,w) &= (\Pi W)_{j+1}^H w_j (\Pi W)_2^{j-1},\\ 
  Q_{j,k}^2(W,w) &= (\Pi W)_{k+1}^H w_k (\Pi W)_{j+1}^{k-1} w_j (\Pi W)_2^{j-1},
\end{align*} 
so that
\begin{align*}
  H_W(w) &= - \sum_{j \neq k \ge 2}^H \tr\left( W_{H+1} Q_{j,k}^2 (W,w) W_{1,1} M^\T \right) - \sum_{j=2}^H \tr \left( w_{H+1} Q_j^1(W,w) W_{1,1} M^\T \right) \\
  & - \sum_{l=2}^H \tr\left( W_{H+1} Q_l^1 (W,w) w_{1,1} M^\T \right) - \tr \left( w_{H+1} (\Pi W)_2^H w_{1,1} M^\T \right) \\ 
  & + \sum_{j \neq k \ge 2}^H \tr \left( W_{H+1} Q_{j,k}^2 (W,w) W_{1,2} W_{1,2}^\T [(\Pi W)_2^{H+1}]^\T \right) \\
  &+ \sum_{j = 2}^H \tr \left( w_{H+1} Q_j^1 (W,w) W_{1,2} W_{1,2}^\T [(\Pi W)_2^{H+1}]^\T \right) \\ 
  &+ \sum_{j = 2}^H \tr \left( W_{H+1} Q_j^1 (W,w) w_{1,2} W_{1,2}^\T [(\Pi W)_2^{H+1}]^\T \right) \\ 
  &+ \tr \left( w_{H+1} (\Pi W)_2^H w_{1,2} W_{1,2}^\T [(\Pi W)_2^{H+1}]^\T \right) \\
  & + \frac12 \left\| \sum_{j=2}^H W_{H+1} Q_j^1 (W,w) W_{1,1} + w_{H+1} (\Pi W)_2^H W_{1,1} + (\Pi W)_2^{H+1} w_{1,1} \right\|_F^2 \\
  & + \frac12 \left\| W_{H+1} \sum_{j=2}^H Q_j^1 (W,w) W_{1,2} + w_{H+1} (\Pi W)_2^H W_{1,2} + (\Pi W)_2^{H+1} w_{1,2} \right\|_F^2.
\end{align*}

The differential and the Hessian of $L$ are given by $\Delta_W(w)$ and $H_W(w)$ respectively. As a consequence, by definition the GDD associated with $L$ is given by 
\begin{equation}\label{eq:deep-gradient}
  \begin{cases}
  \dt{W_{1,1}} &= -\nabla_{W_{1,1}}L = [(\Pi W)_2^{H+1}]^\T M, \\
  \dt{W_{1,2}} &= -\nabla_{W_{1,2}}L=  - [(\Pi W)_2^{H+1}]^\T (\Pi W)_2^{H+1} W_{1,2}, \\
  \dt{W_j} &= -\nabla_{W_j}L=  [(\Pi W)_{j+1}^{H+1}]^\T M W_{1,1}^\T [(\Pi W)_2^{j-1}]^\T \\
  &- [(\Pi W)_{j+1}^{H+1}]^\T (\Pi W)_2^{H+1} W_{1,2} W_{1,2}^\T [(\Pi W)_2^{j-1}]^\T,\quad \text{for $j = 2,\ldots, H$}, \\
  \dt{W_{H+1}} &= -\nabla_{W_{H+1}}L = M W_{1,1}^\T [(\Pi W)_2^H]^\T - (\Pi W)_2^{H+1} W_{1,2} W_{1,2}^\T [(\Pi W)_2^H]^\T.
\end{cases}
\end{equation}
\end{proposition}

\begin{proof}
To proceed, we first expand the non commutative 
product $\Big(\Pi(W+w)\Big)_2^{H}$ to get its linear and quadratic parts (with respect to $w$) by using the notations introduced in the 
proposition,
$$
\big(\Pi(W+w)\big)_2^{H}=(\Pi W)_2^{H}+
\sum_{j=2}^HQ^1_j(W,w)+\sum_{j\neq k\geq 2}Q^2_{j,k}(W,w)
+O \left( \|w\|^3 \right).
$$
We then plug the above equation into the following matrix products
$$
(W_{H+1}+w_{H+1})\big(\Pi(W+w)\big)_2^{H}(W_{1,1}+w_{1,1}),\ \
(W_{H+1}+w_{H+1})\big(\Pi(W+w)\big)_2^{H}(W_{1,2}+w_{1,2}),
$$
that appear in the definition of $L(W+w)$ given in \eqref{eq:deep-loss-function-simple}. For instance, one gets
\begin{align*}
&(W_{H+1}+w_{H+1})\big(\Pi(W+w)\big)_2^{H}(W_{1,1}+w_{1,1})=
(\Pi W)_2^{H+1}W_{1,1}\\
&+w_{H+1}(\Pi W)_2^{H}W_{1,1}+\sum_{j=2}^HW_{H+1}Q^1_j(W,w)W_{1,1}+(\Pi W)_2^{H+1}w_{1,1}\\
&+\sum_{j\neq k\geq 2}W_{H+1}Q^2_{j,k}(W,w)W_{1,1}+
\sum_{j=2}^Hw_{H+1}Q^1_j(W,w)W_{1,1} \\ 
&+\sum_{j=2}^HW_{H+1}Q^1_j(W,w)w_{1,1}+w_{H+1}(\Pi W)_2^{H}w_{1,1}+O \left( \|w\|^3 \right),
\end{align*}
where, on the right-hand side of the equality each line gathers an order of approximation (with respect to $w$). The rest of the computation is straightforward and 
tedious.
\end{proof}

We deduce from the above proposition the following equations verified by  
critical points.
\begin{lemma}\label{lem:crit0}
Let $W$ be a critical point, i.e., an element of $\Crit(L)$. Let $R:= (\Pi W)_2^{H+1}$, 
\begin{equation}\label{eq:deep-critical-condition-3}
  \begin{cases} 
  R^\T S_Y = R^\T R W_{1,1}, \\ 
  R W_{1,2} = 0, \\ 
  R W_{1,1} S_Y = R W_{1,1} W_{1,1}^\T R^\T.
  \end{cases}
\end{equation}
\end{lemma}
\begin{proof}
As a direct consequence of Proposition~\ref{prop:variation-deep}, the set of critical points $\Crit(L)$ is given by
\[
  \nabla_W(w) = 0 \Leftrightarrow 
  \begin{cases} [(\Pi W)_2^{H+1}]^\T M = 0, \\ 
  [(\Pi W)_2^{H+1}]^\T (\Pi W)_2^{H+1} W_{1,2} = 0, \\ 
  [(\Pi W)_{j+1}^{H+1}]^\T M W_{1,1}^\T [(\Pi W)_2^{j-1}]^\T \\- [(\Pi W)_{j+1}^{H+1}]^\T (\Pi W)_2^{H+1} W_{1,2} W_{1,2}^\T [(\Pi W)_2^{j-1}]^\T = 0, \\
  \hfill \text{ for $j = 2,\ldots, H+1$.}
  \end{cases}
\]

By the second equation we have $ (\Pi W)_2^{H+1} W_{1,2} = 0$ and therefore the above equations are reduced to 
\begin{equation}\label{eq:deep-critical-condition-1}
  \begin{cases} 
  [(\Pi W)_2^{H+1}]^\T M = 0, \\ 
  (\Pi W)_2^{H+1} W_{1,2} = 0, \\ 
  [(\Pi W)_{j+1}^{H+1}]^\T M W_{1,1}^\T [(\Pi W)_2^{j-1}]^\T = 0,\ \text{for $j = 2,\ldots, H+1 $}.
  \end{cases} 
\end{equation}
Plugging in the definition $M = S_Y - (\Pi W)_2^{H+1} W_{1,1}$ we obtain
\begin{equation}\label{eq:deep-critical-condition-2}
  \begin{cases} 
  [(\Pi W)_2^{H+1}]^\T S_Y = [(\Pi W)_2^{H+1}]^\T (\Pi W)_2^{H+1} W_{1,1} \\ 
  (\Pi W)_2^{H+1} W_{1,2} = 0 \\ 
  [(\Pi W)_{j+1}^{H+1}]^\T S_Y W_{1,1}^\T [(\Pi W)_2^{j-1}]^\T = [(\Pi W)_{j+1}^{H+1}]^\T (\Pi W)_2^{H+1} W_{1,1} W_{1,1}^\T [(\Pi W)_2^{j-1}]^\T \\
  \hfill \text{for $j = 2,\ldots, H+1 $.}
  \end{cases}
\end{equation}

Note that, with $j=H+1$ of the third equation in \eqref{eq:deep-critical-condition-2} and taking its transpose, we obtain
\begin{equation}\label{eq:deep-relation-j=H+1}
  (\Pi W)_2^H W_{1,1} S_Y = (\Pi W)_2^H W_{1,1} W_{1,1}^\T [(\Pi W)_2^{H+1}]^\T,
\end{equation}
pre-multiplying $W_{H+1}$ on both sides we get
\[
    (\Pi W)_2^{H+1} W_{1,1} S_Y = (\Pi W)_2^{H+1} W_{1,1} W_{1,1}^\T [(\Pi W)_2^{H+1}]^\T.
\]
Using now \eqref{eq:deep-critical-condition-2} we obtain
\eqref{eq:deep-critical-condition-3}.
\end{proof}
 
 This yields the following crucial lemma.
\begin{lemma}[Critical Point conditions in Deep Network]\label{lem:same-ker-deep}
  Assume that Assumptions~\ref{ass:deep-dimension-condition}, 
  ~\ref{ass:full-rank-data} and ~\ref{ass:distinct-singular-values} hold true. For every $W \in \Crit(L)$, define 
  \[
    R = (\Pi W)_2^{H+1}, \quad Z = (\Pi W)_2^H,\quad r = \rank R \in [0,d_y].
  \]
  Then, one has that 
  \begin{description}[leftmargin=!]
    \item[$1)$] $RW_{1,1}=S_Y D_W$ where $D_W \in \mathbb{R}^{d_y \times d_y}$ is a diagonal matrix with $r $ ones and $d_y-r$ zeros on the diagonal. Moreover, $R^\T =R^\T D_W$.
    \item[$2)$] there exists a permutation matrix $U \in \mathbb{R}^{d_y \times d_y}$ such that
    \begin{align*}
      & U^\T W_{H+1} = \begin{bmatrix} W_{H+1,1} \\ W_{H+1,2} \end{bmatrix}, \quad W_{H+1,1} \in \mathbb{R}^{r \times d_H}, \quad W_{H+1,2} \in \mathbb{R}^{ (d_y - r) \times d_H}, \\
      &W_{1,1} U = \left[\begin{array}{c|c} W_{1,1,1} & W_{1,1,2} \end{array}\right], \quad W_{1,1,1} \in \mathbb{R}^{d_1 \times r}, \quad W_{1,1,2} \in \mathbb{R}^{d_1 \times (d_y - r)}, \\
      &W_{H+1,2} Z = 0,\quad W_{H+1,1} Z \in \mathbb{R}^{ r \times d_1} \text{ is of rank } r;
    \end{align*}
  \item[$3)$] if we write 
  \[
    S_Y = \begin{bmatrix} D_Y & 0 \\ 0 & E_Y \end{bmatrix}, \quad D_Y \in \mathbb{R}^{ r \times r}, \quad E_Y \in \mathbb{R}^{ (d_y - r) \times (d_y - r)}, \hbox{ $D_Y$, $ E_Y$ diagonal.}
  \]
  Then Equation~\eqref{eq:deep-critical-condition-3} can be reexpressed as
  \begin{equation}\label{eq:deep-critical-condition-3-simple}
  \begin{cases}
    W_{H+1,1} Z W_{1,1,1} = D_Y, \\
    Z W_{1,1,2} = 0, \\
    W_{H+1,1} Z W_{1,2}  = 0, 
  \end{cases}
  \end{equation}
    \item[$4)$] the critical values of the effective loss function are given by $L(W) = \frac12 \| E_Y \|^2$, i.e., the set of critical values of $L$ is equal to the finite set made of the half sum of the squares of any subset of the singular values of $\bar Y$.
  \end{description} 
\end{lemma}
Note that \eqref{eq:deep-critical-condition-3-simple} is only a  necessary condition of the critical points since we use only the $j=H+1$-th of the total $H$ equations from the last equation of \eqref{eq:deep-critical-condition-2}.

\begin{proof}
Set $Q:=RW_{1,1}$, which is a square $d_y\times d_y$ matrix.
By pre-multiplying the first equation in \eqref{eq:deep-critical-condition-3} with $W_{1,1}^\T$ and rewriting the third equation of \eqref{eq:deep-critical-condition-3}, we obtain
\begin{equation}\label{eq:proof-element}
  Q^\T S_Y = Q^\T Q=S_YQ, \quad QS_Y=QQ^\T=S_YQ^\T
\end{equation}
where we have rendered explicitly the fact that both $Q^\T Q$ and $Q Q^\T$ in the above equation are symmetric, and that $S_Y$ is diagonal. One then deduces that
$$
\begin{cases}
QS_Y^2=(QS_Y)S_Y=Q(Q^\T S_Y)=QQ^\T Q,\\
S_Y^2Q=S_Y(S_YQ)=(S_YQ^\T) Q=QQ^\T Q,
\end{cases}
$$
yielding that $Q$ and $S_Y^2$ commute. By Assumption~\ref{ass:distinct-singular-values}, $S_Y^2$ has non zero and distinct elements, implying that $Q$ is a diagonal matrix. Since $R^\T S_Y = R^\T Q$ according to the first equation in \eqref{eq:deep-critical-condition-3} and $R$ is of rank $r$, it follows that $Q$ has $r$ non zero elements. 
From \eqref{eq:proof-element}, one has that $QS_Y=Q Q^\T=Q^2$ and hence $Q =S_Y D_W$
 for a diagonal matrix $D_W \in \mathbb R^{d_y \times d_y}$ with $r$ ones and otherwise zero on the diagonal. From the first equation of \eqref{eq:deep-critical-condition-3}, it follows that $R^\T = R^\T Q S_Y^{-1} =R^\T D_W$, yielding Item $1)$.

Since $R^\T = R^\T D_W$ with $R \in \mathbb R^{d_y \times d_1}$ and diagonal $D_W \in \mathbb R^{d_y \times d_y}$, there exists a permutation matrix $U \in \mathbb{R}^{d_y \times d_y}$ so that it holds
\begin{equation}
  D_W U = \begin{bmatrix} I_r & 0 \\ 0 & 0_{d_y-r} \end{bmatrix}
\end{equation}
and therefore $R^\T U= \begin{bmatrix} \bar R^\T  & 0_{d_1 \times (d_y - r)} \end{bmatrix}$ for some $\bar{R} \in \mathbb{R}^{r \times d_1}$ of full rank (since $R$ is of rank $r$).
Also, by writing $W_{1,1} U = \left[\begin{array}{c|c} W_{1,1,1} & W_{1,1,2} \end{array}\right]$ with $W_{1,1,1} \in \mathbb{R}^{d_1 \times r}$ and $W_{1,1,2} \in \mathbb{R}^{d_1 \times (d_y - r)}$, then the whole product gives
\[
  U^\T R W_{1,1} U = \begin{bmatrix} \bar R \\ 0 \end{bmatrix} \left[\begin{array}{c|c} W_{1,1,1} & W_{1,1,2} \end{array}\right] = \begin{bmatrix} \bar R W_{1,1,1} & 0 \\ 0 & 0_{d_y - r}, \end{bmatrix}
\]
where we used, for the second equality in the above equation, the fact that $Q: = R W_{1,1}$ is symmetric and so must be $U^\T R W_{1,1} U$.
As such, $ (\bar R W_{1,1,1}) \in \mathbb{R}^{r \times r}$ also symmetric, diagonal, and of full rank (equal to $r$) and $\bar R W_{1,1,2} = 0$.
Note that since $ \bar R $ is of full rank (equal to $r$) and has its $r$ rows linearly independent, we have $W_{1,1,1}$ is also of rank $r$ and therefore the matrix $W_{1,1,1}$ is of minimum rank $r$. 

To prove Item $2)$ of the lemma, it suffices to recall the definition of $R = (\Pi W)_2^{H+1} = W_{H+1} Z$, and to note that
\[
  R^\T U = Z^\T W_{H+1}^\T U = Z^\T \left[\begin{array}{c|c} W_{H+1,1}^\T & W_{H+1,2}^\T \end{array}\right] = \left[\begin{array}{c|c} \bar R^\T & 0 \end{array}\right]
\]
so that
\[
  \begin{cases}
    W_{H+1,1} Z = \bar R\\
    W_{H+1,2} Z = 0.
  \end{cases}
\]
This concludes the proof of Item $2)$ of the lemma.

As a consequence of the change of basis in Lemma~\ref{lem:same-ker-deep}, we rewrite the (necessary) critical conditions \eqref{eq:deep-critical-condition-3} as follows,
\begin{align*}
  &\begin{cases} 
  \left[\begin{array}{c|c} \bar R^\T & 0 \end{array}\right] \left( \begin{bmatrix} D_Y & 0 \\ 0 & E_Y \end{bmatrix} - \begin{bmatrix} \bar R W_{1,1,1} & 0 \\ 0 & 0 \end{bmatrix} \right) = 0\\ 
  \bar R W_{1,2} = 0 \\ 
  \begin{bmatrix} \bar R W_{1,1,1} & 0 \\ 0 & 0 \end{bmatrix} \left( \begin{bmatrix} D_Y & 0 \\ 0 & E_Y \end{bmatrix} - \begin{bmatrix} (\bar R W_{1,1,1})^\T & 0 \\ 0 & 0 \end{bmatrix} \right) = 0
  \end{cases}
  \\
  &\Leftrightarrow
  \begin{cases} 
  \bar R^\T \left( D_Y - \bar R W_{1,1,1} \right) = 0\\ 
  \bar R W_{1,2} = 0 \\ 
  \bar R W_{1,1,1} \left( D_Y - (\bar R W_{1,1,1})^\T \right) = 0
  \end{cases}
\end{align*}
with $S_Y = \begin{bmatrix} D_Y & 0 \\ 0 & E_Y \end{bmatrix}$ and the fact that both $\bar R$ and the product $\bar R W_{1,1,1}$ are of full rank (equal to $r$), we further simplify \eqref{eq:deep-critical-condition-3} as
\[
  \begin{cases}
    D_Y - \bar R W_{1,1,1} = 0 \\
    \bar R W_{1,2}  = 0
  \end{cases}
\]
and conclude by stating that, for $W \in \Crit(L)$, we have
\[
  L(W) = \frac12 \| D_Y - \bar R W_{1,1,1} \|_F^2 + \frac12 \| E_Y \|_F^2 = \frac12 \| E_Y \|_F^2, 
\]
as well as
\begin{equation}\label{eq:M}
  U^\T M U = \begin{bmatrix} 0 & 0 \\ 0 & E_Y \end{bmatrix}.
\end{equation}
Since $U U^\T = I_{d_y}$ for the permutation matrix $U$ introduced above, by post-multiplying \eqref{eq:deep-relation-j=H+1} with $U$ we get
\begin{align*}
  &Z W_{1,1} U U^\T S_Y U = Z W_{1,1} U U^\T W_{1,1}^\T R^\T U \\ 
  &\Leftrightarrow 
  \begin{cases} 
  Z W_{1,1,1} D_Y = Z (W_{1,1,1} W_{1,1,1}^\T + W_{1,1,2} W_{1,1,2}^\T) Z^\T W_{H+1,1}^\T, \\
  Z W_{1,1,2} E_Y = 0,
  \end{cases}
\end{align*}
and therefore $Z W_{1,1,2} = 0$. This concludes the proof of Items $3)$~and~$4)$ of the lemma.
\end{proof}

\subsection{Analysis of the Hessian}
\label{subsec:Hessian-rZ>r}

As discussed in the previous section (in Lemma~\ref{lem:same-ker-deep} particularly), we now have a non trivial description of the set of critical points $\Crit(L)$ that can be written, according to the rank of the matrix product $R = (\Pi W)_2^{H+1}$, as the following disjoint union
\[
  \Crit(L) = \cup_{r=0}^{d_y} \Crit_r(L),
\]
where we denote $\Crit_r(L)$ the set of critical points such that $\rank(R) = r$.

\begin{figure}[htp]
\centering
\includegraphics[width=3in]{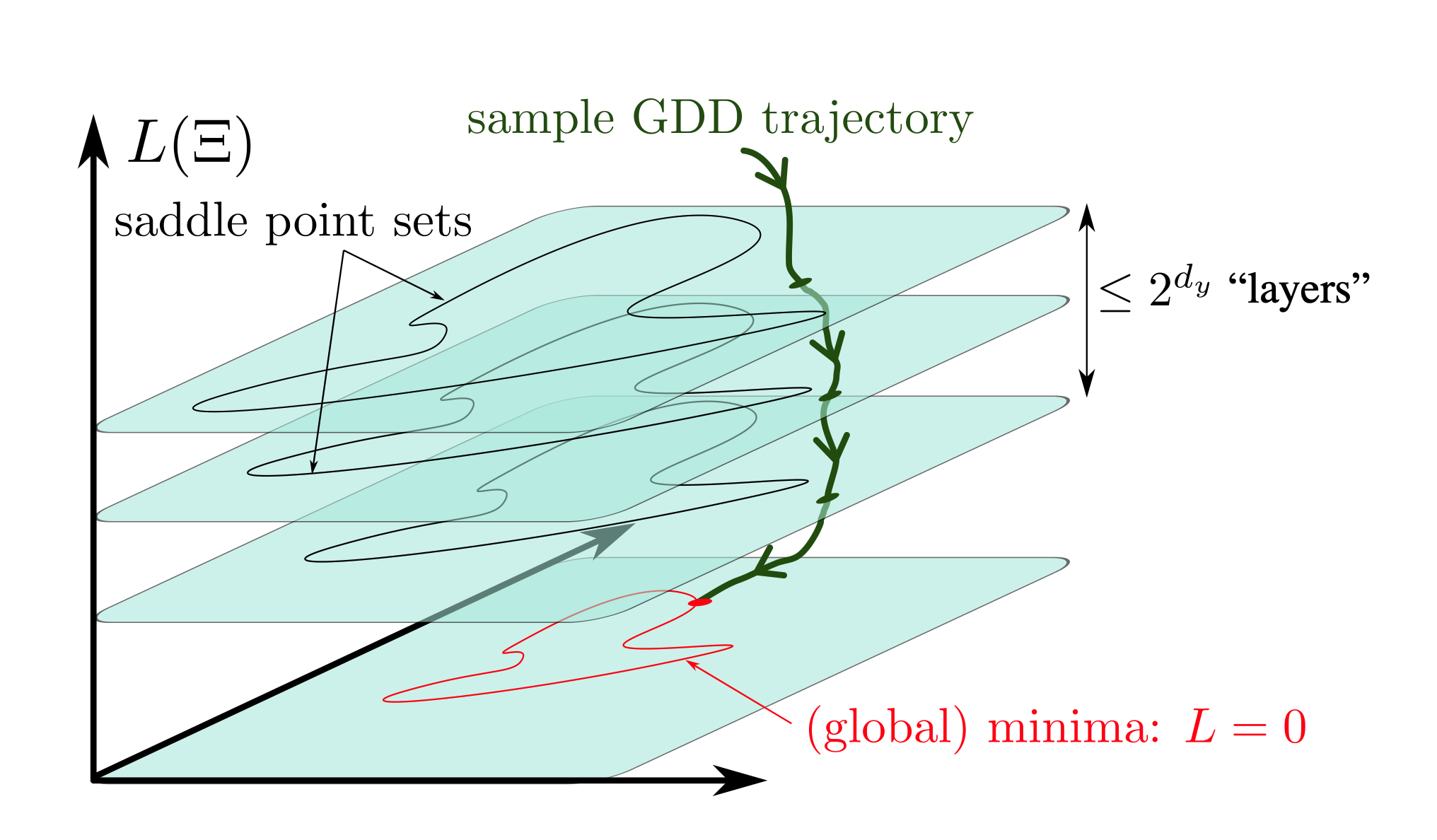}\\
\caption{ { A geometric ``vision'' of the loss landscape. } } 
\label{fig:landscape}
\end{figure}

This description of critical points naturally leads to the following proposition on the loss function \(L(\cdot)\), that can be further ``visualized'' as in Figure~\ref{fig:landscape}.

In order to precisely formulate the next proposition, we recall $r_Z$ the rank of the product $Z:= (\Pi W)_2^H$, which was introduced in Lemma~\ref{lem:same-ker-deep}. 
The next proposition gives results on the landscape of deep linear networks. Note that most of the proposition has been established in \cite[Theorem 2.3, Corollary 2.4]{kawaguchi2016deep}.

\begin{proposition}[Landscape of Deep Linear Networks]
\label{prop:landscape-deep-linear-networks}
Under Assumption~\ref{ass:deep-dimension-condition}-\ref{ass:distinct-singular-values}, for every $W \in \Crit(L)$ the loss function $L(W)$ has following properties:
\begin{description}[leftmargin=!]
\item[$i)$] Every local minimum is a global maximum and every critical point 
that is not a global minimum is a saddle point;
 \item[$ii)$] If $H=1$, the Hessian at every saddle point admits a negative 
 eigenvalue and if $H\geq 2$, there exist saddle points with non negative Hessian;
\item[$iii)$] Every critical point $W \in \Crit_r(L)$ with $0 \le r \le d_y - 1$ is a saddle point. In particular, the set of saddle points is an algebraic variety of positive dimension, i.e., (up to a permutation matrix) the zero set of the polynomial functions given in \eqref{eq:M}, with \(E_Y \neq 0 \). Moreover, if we denote $r_Z$ the rank of the matrix product $Z = (\Pi W)_2^H$ and recall $r:= \rank(\Pi W)_2^{H+1} \le r_Z$. Then if $r_Z > r \ge 0$, the Hessian has at least one negative eigenvalue.
\end{description}
\end{proposition}
\begin{proof}
Items $i)$ and $ii)$ are proved in \cite{kawaguchi2016deep}. Hence we only 
provide a proof of Item $iii)$. For that purpose, we rewrite the associated 
Hessian for all $W \in \Crit_r(L)$ and $w\in\mathcal{X}$, by taking into consideration the second 
equation \eqref{eq:deep-critical-condition-3}
\begin{align*}
  H_W(w) &= - \sum_{j \neq k \ge 2}^H \tr\left( W_{H+1} Q_{j,k}^2 (W,w) W_{1,1} M^\T \right) - \sum_{j=2}^H \tr \left( w_{H+1} Q_j^1(W,w) W_{1,1} M^\T \right), \\
  & - \sum_{j=2}^H \tr\left( W_{H+1} Q_j^1 (W,w) w_{1,1} M^\T \right) - \tr \left( w_{H+1} (\Pi W)_2^H w_{1,1} M^\T \right) 
  \end{align*}
  \begin{align*}
  & + \frac12 \left\| \sum_{j=2}^H W_{H+1} Q_j^1 (W,w) W_{1,1} + w_{H+1} (\Pi W)_2^H W_{1,1} + (\Pi W)_2^{H+1} w_{1,1} \right\|_F^2, \\
  & + \frac12 \left\| W_{H+1} \sum_{j=2}^H Q_j^1 (W,w) W_{1,2} + w_{H+1} (\Pi W)_2^H W_{1,2} + (\Pi W)_2^{H+1} w_{1,2} \right\|_F^2.
\end{align*}
With the change of basis in Lemma~\ref{lem:same-ker-deep} defined by the permutation matrix $U$, the above expression can be further simplified as
\begin{align*}
  &H_W(w) = - \sum_{j \neq k \ge 2}^H \tr \left( Q_{j,k}^2 (W,w) W_{1,1,2} E_Y W_{H+1,2} \right) \\ 
  &- \sum_{j=2}^H \tr \left( Q_j^1(W,w) W_{1,1,2} E_Y w_{H+1,2} \right), \\
  & - \sum_{l=2}^H \tr\left( Q_l^1 (W,w) w_{1,1,2} E_Y W_{H+1,2} \right) - \tr \left( (\Pi W)_2^H w_{1,1,2} E_Y w_{H+1,2} \right) \\ 
  & + \frac12 \left\| \begin{bmatrix} W_{H+1,1} \\ W_{H+1,2} \end{bmatrix} \sum_{j=2}^H Q_j^1 (W,w) \begin{bmatrix} W_{1,1,1} & W_{1,1,2} \end{bmatrix} + \begin{bmatrix} w_{H+1,1} \\ w_{H+1,2} \end{bmatrix} \begin{bmatrix} Z W_{1,1,1} & 0 \end{bmatrix} 
  \right.\\
  &\left.+ \begin{bmatrix} W_{H+1,1} Z \\ 0 \end{bmatrix} \begin{bmatrix} w_{1,1,1} & w_{1,1,2} \end{bmatrix} \right\|_F^2, \\
  & + \frac12 \left\| \begin{bmatrix} W_{H+1,1} \\ W_{H+1,2} \end{bmatrix} \sum_{j=2}^H Q_j^1 (W,w) W_{1,2} + \begin{bmatrix} w_{H+1,1} \\ w_{H+1,2} \end{bmatrix} Z W_{1,2} + \begin{bmatrix} W_{H+1,1} Z \\ 0 \end{bmatrix} w_{1,2} \right\|_F^2,
\end{align*}
where we similarly perform the following decomposition
\begin{align*}
  & U^\T w_{H+1} = \begin{bmatrix} w_{H+1,1} \\ w_{H+1,2} \end{bmatrix}, \quad w_{H+1,1} \in \mathbb{R}^{r \times d_H}, \quad w_{H+1,2} \in \mathbb{R}^{ (d_y - r) \times d_H}, \\
  &w_{1,1} U = \left[\begin{array}{c|c} w_{1,1,1} & w_{1,1,2} \end{array}\right], \quad w_{1,1,1} \in \mathbb{R}^{d_1 \times r}, \quad w_{1,1,2} \in \mathbb{R}^{d_1 \times (d_y - r)},
\end{align*}
and use the fact that $W_{H+1,2} Z = 0$ so that the Hessian becomes a function of $w$ which is given by the coordinates
\[
  w = (w_{1,1,1}, w_{1,1,2}, w_{1,2}, w_2, \ldots, w_H, w_{H+1,1}, w_{H+1,2}).
\] 
 Since $\ker Z \subseteq \ker (W_{H+1,1} Z)$ (both in $\mathbb{R}^{d_1}$), we have 
 \[
   r_Z:= \rank(Z) \ge r = \rank (W_{H+1,1} Z).
 \]
Assume $\rank Z > \rank (W_{H+1,1} Z)$.  We have $\ker Z \subsetneq \ker (W_{H+1,1} Z) $ so that there exists $v \neq 0 \in \ker (W_{H+1,1} Z)$ and $ v \not \in \ker Z$ so that $W_{H+1,1} Z v = 0$ while $V :=Z v \neq 0$.

Then we take $w = (0,\lambda v,0,0,\ldots,0,\mu T)$, with $ \tr (T V E_Y) \neq 0$, i.e., 
\[
  w_{1,1,1} = 0, w_{1,1,2} = \lambda v, w_{1,2} = 0, w_{H+1,1} = 0, w_{H+1,2} = \mu T, w_i = 0,\quad 2 \le i \le H,
\]
with $\lambda$ and $\mu$ are real numbers. Hence, the Hessian becomes a function of $( \lambda, \mu) \in \mathbb{R}^2$, i.e., 
\[
  H_W(\lambda, \mu) = - \tr(T V E_Y) \lambda \mu + \left( \| T Z W_{1,1,1} \|_F^2 +  \| T Z W_{1,2}\|_F^2 \right) \frac{\mu^2}2.
\]
Since $\tr(T V E_Y) \neq 0$, $H_W(w)$ admits at least one negative eigenvalue. 
\end{proof}

\begin{remark}
Recall that Item $i)$ and $ii)$ in Proposition~\ref{prop:landscape-deep-linear-networks} have been previously obtained in 
\cite{kawaguchi2016deep}. However, our findings improve the results of 
\cite{kawaguchi2016deep} in two ways. First of all, our methods are more 
flexible since we only rely on the quadratic form associated with the Hessian 
matrix and we never perform manipulations on the matrix itself, which would 
require handling for example Kronecker products. Secondly, the condition in 
\cite{kawaguchi2016deep} to get a negative eigenvalue for the Hessian 
matrix at a saddle point (Item $(iv)$ in Theorem 2.3 of 
\cite{kawaguchi2016deep}) reads ``$r_Z=\min(d_1,\cdots,d_H)$''. It is easy 
to see that, in that case,  our condition $r_Z>r$ is automatically satisfied 
since one has $r<d_y\leq \min(d_1,\cdots,d_H)$ at a saddle point.
 \end{remark}

\section{On the conjecture (OVF) in the case \texorpdfstring{$H=1$}{H1}}\label{sec:proofOVC}
In this section, we provide a complete argument for the proof of Conjecture (OVF) in the case $H=1$ under the following additional assumption.

\begin{assumption}[Distinct Critical Values for $L$]
The loss function $L$ admits two by two distinct values over two by two distinct subsets made of singular values of the target $\bar Y$ .
\label{ass:distinct-critical-values}
\end{assumption}
Note that the above assumption is stronger than Assumption~\ref{ass:distinct-singular-values} but still it is verified for almost every choice of data-target pair $(X,Y)$.

\bigskip

In the case of a single-hidden-layer $H=1$, we rewrite the gradient system in \eqref{eq:deep-gradient} as
\[
  \begin{cases}
    \dt{W_{1,1}} &= W_2^\T M, \\
    \dt{W_{1,2}} &= - W_2^\T W_2 W_{1,2},\quad W=(W_{1,1},W_{1,2},W_2), \quad M= S_Y - W_2 W_{1,1},\\
    \dt{W_2} &= M W_{1,1}^\T - W_2 W_{1,2} W_{1,2}^\T.
  \end{cases}
\]
The state space is 
$$
\mathcal{X}=\mathbb{R}^{d_1 \times d_y} \times \mathbb{R}^{d_1 \times (d_x - d_y)} \times \mathbb{R}^{d_y \times d_1},
$$
and, at a critical point $W\in \Crit(L)$, we deduce from Lemma~\ref{lem:same-ker-deep} that for $0\leq r\leq d_y-1$,  and up to a change of basis (which belongs to a finite set of orthogonal matrices), the following relations (note here that $Z$ reduces to the $d_H\times d_H$ identity matrix in the case of $H=1$)
\begin{equation}\label{eq:critH1}
  \begin{cases}
    W_{2,1}W_{1,1,1} = D_Y,\quad W_{2,1} \in \mathbb{R}^{r \times d_1}, \  W_{1,1,1} \in \mathbb{R}^{d_1 \times r},\ \rank(W_{2,1})= \rank(W_{1,1,1})=r, \\
    W_{1,1,2} = 0,\quad W_{1,1,2} \in \mathbb{R}^{(d_y-r) \times d_1} \\
    W_{2,1} W_{1,2} = 0,\quad W_{1,2} \in \mathbb{R}^{d_1 \times (d_x - d_y)}, \\
    W_{2,2} = 0,\quad W_{2,2} \in \mathbb{R}^{(d_y-r) \times d_1}.
  \end{cases}
  \end{equation}
  In particular we obtain the following decomposition for $W_{1,1}$ and $W_2$
  \[
W_{1,1} = \begin{bmatrix}W_{1,1,1} &W_{1,1,2} = 0 \end{bmatrix}, \quad W_2 = \begin{bmatrix} W_{2,1} \\ W_{2,2} = 0 \end{bmatrix}.
\]
We also deduce from \eqref{eq:critH1} that $W_2W_{1,2}=0$ yielding the simplified expression for the associated Hessian at $W$
\begin{equation}\label{eq:HessH1-0}
  H_W(w) = -\tr (w_2 w_{1,1} M^\T) + \frac12 \| w_2 W_{1,1} + W_2 w_{1,1} \|_F^2 + \frac12 \| w_2 W_{1,2} + W_2 w_{1,2} \|_F^2.
\end{equation}
\subsection{Computation of dimensions for the critical set}

Using Lemma~\ref{lem:same-ker-deep}, For $0\leq r\leq d_y-1$, we can further stratify $\Crit_r(L)$ according to the value taken by $L$ among the subsets made of singular values of cardinality equal to $r$. Setting $c(r)=C_{d_y}^r$ (where the right-hand-side is a binomial coefficient), one has
\begin{equation}\label{eq:strat-Cr}
\Crit_r(L)=\cup_{l=1}^{c(r)}I_l^r, 
\end{equation}
where, for each subset $S$ made of singular values of cardinality equal to $r$, it corresponds a unique subset $I_l^r$ of  $\Crit_r(L)$ where the value of the loss function is equal to the half sum of the squares of the singular values belonging to $S$. This immediately follows from Assumption~\ref{ass:distinct-critical-values}. It also follows at once that the $I_l^r$'s are two by two distinct.

We have then the following proposition.
\begin{proposition}\label{prop:TX=E0}
For $0\leq r\leq d_y-1$, consider the stratification of $\Crit_r(L)$ defined in \eqref{eq:strat-Cr} and assume that Assumption~\ref{ass:distinct-critical-values} holds true. Then, for $1\leq l\leq c(r)$, the algebraic variety $I_l^r$ is a closed embedded (differential) submanifold of $\mathcal{X}$ of dimension $d(r)$ given by
\begin{equation}\label{eq:dimTI}
d(r)=rd_1+(d_y-r)r+(d_y-r)(d_x-d_y).
\end{equation}
Moreover, at a critical point $W$ of $ I_l^r$, the tangent space to $I_l^r$ at $W$ is equal to the subspace corresponding to the zero eigenvalues of the Hessian of $L$ at $W$.
In particular, one has the orthogonal decomposition 
\begin{equation}
\label{eq:dim-splitting}
\mathcal{X}=E^+(W)\oplus T_{W}I_l^r\oplus E^{-}(W),
\end{equation}
where $E^+(\bar W)$ (resp. $E^{-}(\bar W)$) is the eigenspace of $H_{\bar W}$ associated with positive (resp. negative) eigenvalues.
\end{proposition}

\begin{proof}
From now on, fix $0\leq r\leq d_y-1$ and $1\leq l\leq c(r)$. 

Let $\bar W$ be a critical point in $I_l^r$, by performing a SVD on $\bar W_{2,1}$ we obtain
\[
  \bar W_{2,1} = \begin{bmatrix} A & 0 \end{bmatrix} V,
\]
with $A\in \mathbb{R}^{r \times r}$ invertible.
We assume in the sequel that  $r>0$ and leave the special case of $r=0$ to Remark~\ref{rem:r=0} below.

We further decompose the other $\bar W$'s as follows
\begin{align*}
V \bar W_{1,1,1}&= \begin{bmatrix} B_1 \\ B_2 \end{bmatrix},\ 
B_1 \in \mathbb{R}^{r \times r},\ B_2 \in \mathbb{R}^{(d_1 - r) \times r}\\
V \bar W_{1,2}&=  \begin{bmatrix} C_1 \\ C_2 \end{bmatrix},\ 
C_1 \in \mathbb{R}^{r \times (d_x-d_y)},\ C_2 \in \mathbb{R}^{(d_1 - r) \times (d_x - d_y)}.
 \end{align*}
 so that
\begin{align*}
  \bar W_{2,1} \bar W_{1,1,1} = D_Y &\Leftrightarrow \begin{bmatrix} A & 0 \end{bmatrix} \begin{bmatrix} B_1 \\ B_2 \end{bmatrix} = A B_1 = D_Y \Leftrightarrow B_1 = A^{-1} D_Y, \\
  \bar W_{2,1} \bar W_{1,2} = 0 &\Leftrightarrow \begin{bmatrix} A & 0 \end{bmatrix} \begin{bmatrix} C_1 \\ C_2 \end{bmatrix} = A C_1 = 0 \Leftrightarrow C_1 = 0.
\end{align*}
We now consider first order variations around $\bar W$ and we set 
\[
  w_{2,1} V^\T = \begin{bmatrix} a_1 & a_2 \end{bmatrix}, \quad V w_{1,1,1} = \begin{bmatrix} b_1 \\ b_2 \end{bmatrix}, \quad V w_{1,2} = \begin{bmatrix} c_1 \\ c_2 \end{bmatrix}, \quad V w_{1,1,2} = \begin{bmatrix} d_1 \\ d_2 \end{bmatrix},
\]
and $w_{2,2} V^\T = \begin{bmatrix} e_1 & e_2 \end{bmatrix}$.

To obtain the equations of tangent vectors, it is enough to differentiate \eqref{eq:critH1} to obtain
\begin{equation}\label{eq:critH1-1}
  \begin{cases}
    A b_1 + a_1 A^{-1} D_Y + a_2 B_2 = 0, \\
    d_1 = 0, \quad d_2 = 0, \\
    A c_1 + a_2 C_2 = 0, \\
    e_1 = 0, \quad e_2 = 0.
  \end{cases}
\end{equation}
We perform the following linear change of variables
\begin{align*}
  \beta_1&:=Ab_1+a_1B_1+a_2B_2,\ \gamma_1:=Ac_1+a_2C_2,\ 
  \delta_1:=Ad_1, \\ 
  \epsilon_1^\T&:=e_1 B_1 + e_2 B_2, \ 
  \delta_2:=d_2-B_2D_Y^{-1} A d_1.
\end{align*}
We deduce that \eqref{eq:critH1-1} reduces to 
\begin{equation}\label{eq:critH1-2}
\beta_1=0,\ \gamma_1=0,\ \delta_1=\delta_2=0,\ \epsilon_1=e_2=0.
\end{equation}
As such, we get that there is no constraint on the variations $a_1,a_2,b_2, c_2$ and we obtain that the above equation define a linear subspace in $\mathcal{X}$ of dimension $d(r)$ as defined in \eqref{eq:dimTI}. Since this dimension is independent of $\bar W$ (and also of $l$), one deduces that $I_l^r$ is an immersed submanifold of $\mathcal{X}$ of dimension $d(r)$.
Moreover, it is easy to get that any small enough neighborhood of $\bar W$ in $I_l^r\subset \mathcal{X}$ (up to a change of variables only depending on $\bar W$) is the image of a neighborhood of  the origin in $\mathbb{R}^{d(r)}$ by the mapping
\begin{align*}
\mathbb{R}^{d(r)}&\to \mathcal{X}\\
(a_1,a_2,b_2, c_2)&\mapsto (A+a_1,a_2,B_1+b_1,B_2+b_2,c_1,C_2+c_2,0,0,0,0),
\end{align*}
with $b_1=-(A+a_1)^{-1}(a_1B_1+a_2B_2+a_2b_2)$ and $c_1=-(A+a_1)^{-1}(Ac_1+a_2C_2+a_2c_2)$. One deduces that the inclusion map corresponding to $I_l^r\subset \mathcal{X}$ is closed. Hence $ I_l^r$ is an embedded submanifold of $\mathcal{X}$, which is also a closed subset of $\mathcal{X}$.

We next prove the second part of the proposition. Using the previous notations for the variations, we first simplify $H_{\bar W}$, the Hessian at $\bar W$, as follows,
\begin{align*}
  &H_{\bar W} = - \tr( w_{1,1,2} E_Y w_{2,2} ) + \frac12 \left\| \begin{bmatrix} w_{2,1} \bar W_{1,1,1} + \bar W_{2,1} w_{1,1,1} & \bar W_{2,1} w_{1,1,2} \\ w_{2,2} \bar W_{1,1,1} & 0 \end{bmatrix} \right\|_F^2\\
  & + \frac12 \left\| \begin{bmatrix}  w_{2,1} \bar W_{1,2} + \bar W_{2,1} w_{1,2} \\ w_{2,2} \bar W_{1,2} \end{bmatrix} \right\|_F^2 
  = - \tr( E_Y e_1 d_1) - \tr (E_Y e_2 d_2) \\ 
  &+ \frac12 \| a_1 A^{-1} D_Y + a_2 B_2 + A b_1 \|_F^2\\
  & + \frac12 \| A d_1 \|_F^2 + \frac12 \| e_1 A^{-1} D_Y + e_2 B_2 \|_F^2 
  + \frac12 \| a_2 C_2 + A c_1 \|_F^2 + \frac12 \| e_2 C_2 \|_F^2.
\end{align*}
With the change of variable in \eqref{eq:critH1-2}, the Hessian further simplifies to
\begin{equation}\label{eq:HessH1_2}
  H_{\bar W}=\frac12\big(\Vert \beta_1\Vert_F^2+\Vert \gamma_1\Vert_F^2+\Vert \delta_1\Vert_F^2+\Vert \epsilon_1\Vert_F^2+\Vert e_2 C_2\Vert_F^2\big)- \tr( E_Y \epsilon_1^\T D_Y^{-1} \delta_1)- \tr (E_Y e_2 \delta_2).
\end{equation}
We denote $T_{\bar W}I_l^r$ the tangent space of $I_l^r$ at $\bar W$. Let us show next that the restriction of $H_{\bar W}$ to the orthogonal of $T_{\bar W}I_l^r$ (in $\mathcal{X}$) has non zero eigenvalues. The latter space is equal to the points 
where the coordinates $a_1,a_2,b_2,c_2$ are all zero, i.e., the subspace corresponding to any variation $(a_1 = 0,a_2 = 0,\beta_1,b_2 = 0,\gamma_1, c_2 =0,\delta_1,\delta_2,\epsilon_1,e_2)$. To prove this, it suffices to consider the following two quadratic forms
\begin{equation}
\begin{cases}
Q_1(\delta_1,\epsilon_1)=\frac12\big(\Vert \delta_1\Vert_F^2+\Vert \epsilon_1 \Vert_F^2\big)- \tr( E_Y \epsilon_1^\T D_Y^{-1}\delta_1),\\
Q_2(\delta_2,e_2)=\frac12\Vert e_2 C_2\Vert_F^2- \tr (E_Y e_2 \delta_2),
\end{cases}
\end{equation}
since 
\[
  H_{\bar W}-Q_1(\delta_1,\epsilon_1)-Q_2(\delta_2,e_2)=\frac12\big(\Vert \beta_1\Vert_F^2+\Vert \gamma_1\Vert_F^2\big)
\]
only provides positive eigenvalues.

Let us start by considering $Q_1$. Note that both $\delta_1$ and $\epsilon_1$ belong to 
$\mathbb{R}^{r\times (d_y-r)}$. By expressing $Q_1$ with the coefficients of $\delta_1$ and $\epsilon_1$  and by taking account that $D_Y$ and $E_Y$ are diagonal, one deduces that $Q_1$ is the sum of $r(d_y-r)$ quadratic forms over $\mathbb{R}^2$ of the type
$$
Q_1^{i,j}(x,y)=\frac12(x^2+y^2)-\frac{[E_Y]_{jj}}{[D_Y]_{ii}}xy,\quad 1\leq i\leq r,\ 
1\leq j\leq d_y-r.
$$
Thanks to Assumption~\ref{ass:distinct-singular-values}, we deduce that each $Q_1^{i,j}$
has either two positive eigenvalues or one positive and one negative eigenvalue (depending whether $\frac{[E_Y]_{jj}}{[D_Y]_{ii}}<1$ or not).

\medskip

For the sake of studying $Q_2$, we consider $K :=C_2 C_2^\T \in \mathbb{R}^{(d_1 -r) \times (d_1 -r)}$. We have that $K=U_K^\T D_KU_K$ where $U_K$ is an orthogonal matrix and 
$D_K$ is diagonal with non negative elements $(\alpha_1,\cdots,\alpha_{d_1-r})$.
Denoting $\epsilon_2=U_K e_2^\T$ and $\bar\delta_2=U_K\delta_2$, one deduces the following expression for $Q_2$,
\[
  Q_2(\bar\delta_2,\epsilon_2)=\frac12 \tr(\epsilon_2^\T D_K\epsilon_2)- \tr (E_Y \epsilon_2^\T \bar\delta_2).
\]
By expressing $Q_2$ with the coefficients of $\bar\delta_2, \epsilon_2 \in \mathbb{R}^{(d_1 - r) \times (d_y - r)}$ and by taking account that $E_Y$ is diagonal, one deduces that $Q_2$ is the sum of $(d_y-r)^2$ quadratic forms over $\mathbb{R}^2$ of the type
\[
  Q_2^{i,j}(x,y) = \frac{\alpha_i}2x^2-[E_Y]_{jj}xy, \quad 1 \le i,j \le d_y -r.
\]
It is immediate to see that such a quadratic form admits one positive and one negative 
eigenvalue, regardless of the fact that $\alpha_i>0$ or not.
\end{proof}

\begin{remark}\label{rem:r=0}
 In the case where $r=0$, \eqref{eq:critH1-1}  and \eqref{eq:HessH1_2} reduce to 
\[
  d_2=e_2=0,\hbox{ and } H_{\bar W}= - \tr (E_Y e_2 \delta_2) + \frac12 (\| \epsilon_1 \|_F^2 + \| e_2 C_2 \|_F^2)
\]
respectively, which, following the same line of arguments above, yields the statement of Proposition~\ref{prop:TX=E0}.
\end{remark}

\subsection{On the conjecture (OVF) in the case $H=1$}

\subsubsection{Normal Hyperbolicity}
The key notion that enables us to prove the conjecture (OVF) is the of \emph{normal hyperbolicity} \cite{HPS,Pesin} and we recall next this key notion and apply it to the gradient system under consideration.

\begin{definition}\label{def:NH0}
Let $M$ be a Riemannian manifold of dimension $m$, with associated norm $\Vert\cdot\Vert$ on $TM$, the tangent bundle of $M$. A diffeomorphism $f$ of $M$ is said to be normally hyperbolic along a compact submanifold $N$ of dimension $n$, if $N$ is invariant under $f$ and the tangent bundle of $M$ along $N$ has a splitting $T_zM= E^s(z)\oplus T_zN\oplus E^u(z)$, $z\in N$, such that $df E^{v}(z)=E^{v}(f(z))$, with $v\in\{s,u\}$, i.e., $f$ preserves the splitting, and there exits $\lambda_1\leq \mu_1<\lambda_2\leq \mu_2<\lambda_3\leq \mu_3$ with $\mu_1<1<\lambda_3$, such that
\[
  \lambda_1\leq \Vert df_{\vert E^s}\Vert\leq \mu_1,\ \
\lambda_2\leq \Vert df_{\vert TN}\Vert\leq \mu_2,\ \
\lambda_3\leq \Vert df_{\vert E^u}\Vert\leq \mu_3.
\]
\end{definition}
We denote $E^s$ and $E^u$ the distributions on $N$ defined by the mappings $z\mapsto E^s(z)$ and $z\mapsto E^u(z)$, respectively. In particular, they have constant rank, denoted $m_s$ and $m_u$ respectively. 

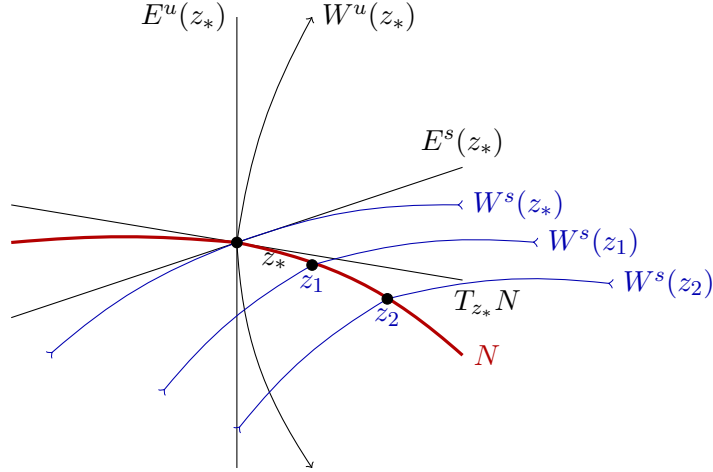
\begin{figure}[htb]
\centering
\begin{tikzpicture}
  \draw
  (3,-.5) coordinate (EX1) node[below,xshift=0.02\textwidth] {$T_{z_*}N$}
  -- (0,0) coordinate (x0) node[below,xshift=0.03\textwidth] {$z_*$}
  -- (-3,.5) coordinate (EX2) node {};
  \draw 
  (0,3) coordinate (Eu1) node[left] {$E^u(z_*)$}
  -- (0,0) -- (0,-3) coordinate (Eu2) node {};
  \draw 
  (-3,-1) coordinate (Es1) node {}
  -- (0,0) -- (3,1) coordinate (Es2) node[above] {$E^s(z_*)$};
  \draw[<-]
  (1,3) coordinate (Wu1) node[right] {$W^u(z_*)$}
  to [bend right=10] (0,0) node {};
  \draw[->]
  (0,0) node {} to [bend right=15] (1,-3) coordinate (Wu2) node {};
  \draw[>-,color=BLUE]
  (3,.5) coordinate (Ws1) node[right] {$W^s(z_*)$}
  to [bend right=10] (0,0) node {};
  \draw[-<,color=BLUE]
  (0,0) node {} to [bend right=10] (-2.5,-1.5) coordinate (Ws2) node {};
  \draw[very thick,color=RED]
  (3,.-1.5) coordinate (X1) node[right] {$N$}
  to [bend right=15] (0,0) node {};
  \draw[very thick,color=RED]
  (0,0) node {} to [bend right=5] (-3,0) coordinate (X2) node {};
  \draw[>-,color=BLUE]
  (4,0) coordinate (Ws1) node[right] {$W^s(z_1)$}
  to [bend right=10] (1,-.3) node[below] {$z_1$};
  \draw[-<,color=BLUE]
  (1,-.3) node {} to [bend right=10] (-1,-2) coordinate (Ws2) node {};
  \draw[>-,color=BLUE]
  (5,-.55) coordinate (Ws1) node[right] {$W^s(z_2)$}
  to [bend right=10] (2,-.75) node[below] {$z_2$};
  \draw[-<,color=BLUE]
  (2,-.75) node {} to [bend right=10] (0,-2.5) coordinate (Ws2) node {};
  \filldraw (0,0) circle (2pt);
  \filldraw (1,-.3) circle (2pt);
  \filldraw (2,-.75) circle (2pt);
\end{tikzpicture}
\caption{Visual representation of normal hyperbolicity.}
\label{fig:normal-hyperbolicity-3D}
\end{figure}

The above property essentially says that the contraction (resp. expansion) effect induced by $f$ in the the stable (resp. unstable) direction $E^s$ (resp. $E^u$) is stronger than the effect of $f$ tangentially to $N$. One can show that $E^s$ and $E^u$ are locally integrable and then construct the local stable and unstable manifolds, $W^s(z)$ and $W^u(z)$ respectively tangent to $E^s(z)$ and $E^u(z)$ at each point $z\in N$. Also, define 
\[
  W^{sn}=\cup_{z\in N}W^s(z),\quad W^{un}=\cup_{z\in N}W^u(z),
\]
the local stable (resp. unstable) manifold of $N$, cf. Figure~\ref{fig:normal-hyperbolicity-3D}. We have the following theorem (cf. \cite[Theorem 3.5]{HPS} and also \cite{Pesin}) that provides fundamental information on $W^s(z)$ and $W^u(z)$. 

\begin{theorem}[Hirsh-Pugh-Shub]\label{the:regNH}
Assume that the hypotheses on the diffeomorphism $f$ are satisfied. Then the local stable and unstable manifolds of $f$, $W^{sn}$ and $W^{un}$, are differential manifolds of class at least $C^1$ of dimension $m_s+n=m-m_u$ and $m_u+n$ respectively.
\end{theorem}

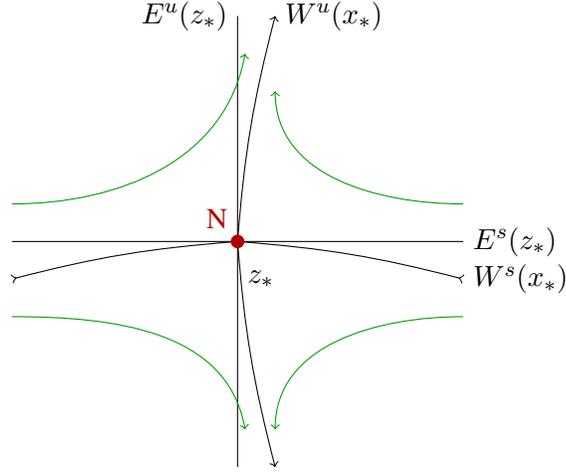
\begin{figure}[htb]
\centering
\begin{tikzpicture}
  \draw
  (3,0) coordinate (Es1) node[right] {$E^s(z_*)$}
  -- (0,0) coordinate (x0) node[right,yshift=-0.03\textwidth] {$z_*$}
  -- (-3,0) coordinate (Es2) node {};
  \draw 
  (0,3) coordinate (Eu1) node[left] {$E^u(z_*)$}
  -- (0,0) -- (0,-3) coordinate (Eu2) node {};
  \draw[<-]
  (.5,3) coordinate (Wu1) node[right] {$W^u(x_*)$} to [bend right=5] (0,0) node {};
  \draw[->]
  (0,0) node {} to [bend right=5] (.5,-3) coordinate (Wu2) node {};
  \draw[>-](3,-.5) coordinate (Ws1) node[right] {$W^s(x_*)$} to [bend right=5] (0,0) node {};
  \draw[-<](0,0) node {} to [bend right=5] (-3,-.5) coordinate (Ws2) node {};
  \draw[->,GREEN](3,.5) coordinate (t11) node {} to [out=180,in=270] (.5,2) coordinate (t12) node {};
  \draw[->,GREEN](-3,.5) coordinate (t21) node {} to [out=0,in=260] (.1,2.5) coordinate (t22) node {};
  \draw[->,GREEN](3,-1) coordinate (t11) node {} to [out=180,in=90] (.5,-2.5) coordinate (t12) node {};
  \draw[->,GREEN](-3,-1) coordinate (t21) node {} to [out=0,in=100] (.1,-2.5) coordinate (t22) node {};
  \draw (0,0) node[circle,fill=RED,scale=0.5] {} ;
  \draw (0,0.3) node[left] {\RED N} ;
\end{tikzpicture}
\caption{Visual representation of normal hyperbolicity and trajectories samples (in {\GREEN \textbf{green}}).}
\label{fig:normal-hyperbolicity-2D}
\end{figure}

\subsubsection{Proof of the conjecture (OVF) in the case $H=1$.}
On the basis of Proposition~\ref{prop:TX=E0}, we are now in place to complete the proof of the conjecture (OVF) in the case $H=1$. We first order the differential manifolds $I_l^r$, for $0\leq r\leq d_y-1$ and $1\leq l\leq c(r)$, according to decreasing values of $L$ and relabel them $I(j)$, for  $1\leq j\leq 2^{d_y}$. We label in accordance the critical values of $L$ by $L(j)$, for  $1\leq j\leq 2^{d_y}$.
Hence, $I(1)$ is equal to $I_1^0$, $L(1)=\frac12\Vert S_Y\Vert_F^2$ and  $I(2^{d_y})$ is equal to the set of global minima with $L(2^{d_y})=0$.

Let us fix $1\leq j\leq 2^{d_y}$. We will next apply repeatedly 
Theorem~\ref{the:regNH}. The Riemannian manifold $M$ is here equal to $
\mathcal{X}$ equipped with the Frobenius norm. The diffeomorphisms $f$  
will be $\Phi_{GDD}^T$, the flows of GDD in time $T>0$ small enough (and 
which can change possibly). As for the compact manifolds $N$, they will be 
compact neighborhoods $N_{w}$ where $w$ in $I(j)$. Moreover,  since $I(j)$ is an invariant closed embedded submanifold of $\mathcal{X}$ and by taking into account Proposition~\ref{prop:TX=E0}, there exists for every $w\in I(j)$ a bounded neighborhood $N_{w}$ of $w$ in $I(j)$ and $T>0$ small enough such that $\Phi_{GDD}^T$ is normally hyperbolic along $N_w$. Indeed, for every $w'\in N_w$, $N_w$ is invariant by $\Phi_{GDD}^T$ and \eqref{eq:dim-splitting} provides the appropriate splitting of $T_{w'}N_w$.

We apply Theorem~\ref{the:regNH} to deduce that $W^{sn}(w)$ and $W^{un}(w)$ are differential proper submanifolds of $\mathcal{X}$ of class $C^1$ since both $m_s$ and $m_u$ are positive according to Proposition~\ref{prop:TX=E0}. In particular, the dimension of $W^{sn}(w)$ is strictly less than that of $\mathcal{X}$. Since $L$ is strictly decreasing along trajectories lying outside $I(j)$, one has that $L>L(j)$ on $W^{sn}(w)\setminus N_{w}$ and $L<L(j)$ on $W^{un}(w)\setminus N_{w}$.  
One can also associate with $w$  an open neighborhood $O_{w}$ of $w$ in 
$\mathcal{X}$ small enough such that, along every trajectory starting in  $O_{w}\setminus (O_{w}\cap W^{sn}(w))$, the value of $L$ becomes smaller than $L(j)$. Indeed, every such a trajectory will approach $O_{w}\cap W^{un}(w)$ at an exponential rate, cf. Figure~\ref{fig:normal-hyperbolicity-2D}. 

We can now conclude the proof of the conjecture (OVF) in the case \texorpdfstring{$H=1$}{H1}. We argue by contradiction and assume that there exists a subset $G_0$ of $\mathcal{X}$ with positive measure such that every trajectory of GDD starting in $G_0$ converges to a saddle point. Pick a point $g_0\in G_0$ such that the trajectory of GDD starting at $g_0$ converges to some $w_{g_0}$ belonging to some $I(j_0)$, with $j_0<2^{d_y}$.
By eventually shrinking $G_0$, we can assume that $(a)$
the infimum value of $L$ on $G_0$ is larger than $L(j_0)$ and $(b)$ there exists a positive time $T_0$ such that $\Phi_{GDD}^{T_0}(G_0)$ is contained in 
$O_{w_{g_0}}\setminus (O_{w_{g_0}}\cap W^{sn}(w_{g_0}))$. By again eventually shrinking $G_0$, there exists a positive time $T_1$ such that the supremum value of $L$ on 
$G_1:=\Phi_{GDD}^{T_1+T_0}(G_0)$
 is smaller than $L(j_0)$. Note that $G_1$ has positive measure. Pick a point $g_1\in G_1$ which converges to some $w_{g_1}$ belonging to some $I(j_1)$, with $j_1<2^{d_y}$. Since $L$ is decreasing along (GDD), one deduces that $j_1>j_0$.

We can now iterate the construction that enabled us to pass from $G_0$ and $g_0$ to $G_1$ and $g_1$. We hence build a sequence of sets $G_p$, $p\geq 0$ of positive measure and a sequence of integers $j_p$ with $1\leq j_p\leq 2^{d_y}$. Since this sequence is increasing, there exists $p_*\geq 1$ such that $j_{p_*}=2^{d_y}$ and hence the trajectories of (GDD) starting in $G_{p_*}$ must converge to global minima. By construction, this implies that there exists a subset $G'_0$ of $G_0$ of positive measure such that the trajectories of (GDD) which start in $G'_0$ converge to global minima. This contradicts the definition of $G_0$ and thus concludes the proof of the conjecture (OVF) in the case $H=1$.

\section{Conclusion}

In this paper, we address the issue of \emph{global} behavior of the gradient descent dynamics in linear neural networks. That behavior is fully characterized, in the sense that, with an intrinsic structural property of the (cascading) network (Lemma~\ref{lem:invariant-in-deep-GDD}), we show a \emph{global} convergence to critical points of all trajectories of the gradient flow via Lojasiewicz's theorem, which helps eliminate the possibility of divergence and even directly establish exponential rate convergence for specific  initializations. Then with a fine local study of critical points we exclude the (possible) worries concerning the ``accumulation'' of saddle points together with associated basin of attractions so that they form ``disjoint layers'' that are of total measure zero in the total weight space. Our results need no unrealistic assumptions for example the (a prior) bound on the Hessians of all critical points, or the network width to grow polynomially with respect to its depth, thereby shed new light on the behavior of simple gradient descent method in the elaborate but particular system of deep neural networks.

When nonlinear networks are considered, by exploring a random model setting for $(X,Y)$, the authors in \cite{choromanska2015loss} argue that the loss surfaces of these networks loosely recall (yet is formally quite different from) a spin-glass model, familiar to statistical physicists. In this case, as the network gets large, local minima gather in a thin ``band'' of similar losses isolated from the global minimum. Stating that the number of local minima outside that band diminishes exponentially with the size of the network, the authors argue that the gradient descent dynamics (in their case the stochastic gradient descent dynamics) converges to this band and therefore leads to deep \emph{nonlinear} networks with good generalization performance. Taking advantage of a random nature for $(X,Y)$ in our present setting would allow for a refinement of our proposed geometric vision, likely by means of a ``statistical extension'' of the key Lemma~\ref{lem:invariant-in-deep-GDD}.

Most discussions on the landscape of deep linear networks (e.g., all local minima are global) are restricted to square loss functions \cite{baldi1989neural,kawaguchi2016deep} for simplicity. However, similar results can be obtain for more general convex differentiable losses \cite{laurent2018deep}. It would be thus of interest to extend the present results to more general objective functions, as well as various optimization methods that are of more practical interest as discussed in Remark~\ref{rem:extension-lojasiewicz}.

\section*{Acknowledgments}

The authors would like to warmly thank the anonymous reviewers for their precise, numerous, and construct comments that dramatically improved this paper and also J. B. Caillau for his helpful discussions on normal hyperbolicity.  

The work of YC is supported by a public grant overseen by the French National Research Agency (ANR) as part of the ``Investissement d'Avenir'' program, through the iCODE project funded by the IDEX Paris-Saclay, ANR-11-IDEX-0003-02. 
ZL would like to acknowledge the National Natural Science Foundation of China (NSFC-12141107), the Fundamental Research Funds for the Central Universities of China (2021XXJS110), the Key Research and Development Program of Hubei (2021BAA037), and the CCF-Hikvision Open Fund (20210008) for providing partial support.
RC would like to acknowledge the MIAI LargeDATA chair (ANR-19-P3IA-0003) at University Grenobles-Alpes as well as the HUAWEI LarDist project for providing partial support of this work.









\bibliography{liao}
\bibliographystyle{plain}

\medskip
Received 17 Apr 2021; revised 10 Jan 2022; early access xxxx 20xx.
\medskip

\end{document}